\documentclass[twoside,11pt]{article}
\usepackage{blindtext}
\usepackage{jmlr2e}

\usepackage{amsmath,amssymb,float,graphicx,algorithm2e,subcaption,bm, multirow,array}
\usepackage{tikz}
\usepackage{epsf,epsfig,pstricks,pst-node,color}
\usepackage{subcaption}

\DeclareMathOperator*{\argmax}{arg\,max}

\newcommand{\matB}{\mathbf{B}}
\newcommand{\matT}{\mathbf{T}}
\newcommand{\matU}{\mathbf{U}}
\newcommand{\matV}{\mathbf{V}}
\newcommand{\matW}{\mathbf{W}}
\newcommand{\veca}{\mathbf{a}}
\newcommand{\vecb}{\mathbf{b}}
\newcommand{\vecc}{\mathbf{c}}
\newcommand{\vecv}{\mathbf{v}}
\newcommand{\vecy}{\mathbf{y}}
\newcommand{\vecnu}{\mathbf{\nu}}
\newcommand{\indicator}{\mathbb{I}}
\newcommand{\R}{\mathbb{R}}
\newcommand{\loss}{\mathcal{L}}
\newcommand{\E}{\mathbb{E}}
\newcommand{\transpose}{\top}
\DeclareMathOperator*{\sign}{sign}
\DeclareMathOperator{\bernoulli}{Bernoulli}

\newcommand{\smallhspace}{\hspace{1pt}}

\newcommand{\repeatthanks}{\footnotemark[1]}
\makeatletter
\renewcommand*{\@makefntext}[1]{{\@setpar{\@@par\@tempdima \hsize 
	\advance\@tempdima-15pt\parshape \@ne 15pt \@tempdima}\par
	\parindent 2em\noindent \hbox to \z@{\hss{\@thefnmark} \hfil}#1}
}
\makeatother

\newcommand{\D}{\mathcal{D}}  
\newcommand{\kl}{\mbox{KL}}

\newcommand{\vecx}{\mathbf{x}}

\def\reg{{\rm\ooalign{\hfil
     \raise.07ex\hbox{\scriptsize R}\hfil\crcr\mathhexbox20D}}}

\usepackage{lastpage}
\jmlrheading{25}{2024}{1-\pageref{LastPage}}{8/18; Revised
12/23}{2/24}{18-566}{Wolfgang Roth, G\"unther Schindler, Bernhard Klein, Robert Peharz, Sebastian Tschiatschek, Holger Fr\"oning, Franz Pernkopf and Zoubin Ghahramani}

\ShortHeadings{Resource-Efficient Neural Networks for Embedded Systems}{Roth, Schindler, Klein, Peharz, Tschiatschek, Fr\"oning, Pernkopf and Ghahramani}

\begin{document}

\title{Resource-Efficient Neural Networks for Embedded Systems} 

\author{\name Wolfgang~Roth\thanks{These authors contributed equally.} \email roth@tugraz.at\\
\addr Graz University of Technology, Austria\\ Laboratory of Signal Processing and Speech Communication
\AND
\name G\"unther Schindler\repeatthanks \email guenther.schindler@ziti.uni-heidelberg.de\\
\addr Heidelberg University, Germany\\ Institute of Computer Engineering
\AND
\name Bernhard Klein\repeatthanks \email bernhard.klein@ziti.uni-heidelberg.de\\
\addr Heidelberg University, Germany\\ Institute of Computer Engineering
\AND
\name Robert Peharz \email robert.peharz@tugraz.at\\
\addr Graz University of Technology, Austria\\  Institute of Theoretical Computer Science
\AND
\name Sebastian Tschiatschek \email sebastian.tschiatschek@univie.ac.at\\
\addr University of Vienna, Austria\\ Faculty of Computer Science
\AND
\name Holger Fr\"oning \email holger.froening@ziti.uni-heidelberg.de\\
\addr Heidelberg University, Germany\\ Institute of Computer Engineering
\AND
\name Franz Pernkopf \email pernkopf@tugraz.at\\
\addr Graz University of Technology, Austria\\ Laboratory of Signal Processing and Speech Communication
\AND
\name Zoubin Ghahramani \email zoubin@eng.cam.ac.uk\\
\addr University of Cambridge, UK
}

\editor{Russ Greiner} 

\maketitle

\makeatletter
\renewcommand*{\@makefntext}[1]{{\@setpar{\@@par\@tempdima \hsize 
             \advance\@tempdima-15pt\parshape \@ne 15pt \@tempdima}\par
             \parindent 2em\noindent \hbox to \z@{\hss{\@thefnmark}. \hfil}#1}
}
\makeatother

\begin{abstract}%
While machine learning is traditionally a resource intensive task, embedded systems, autonomous navigation, and the vision of the Internet of Things fuel the interest in resource-efficient approaches.
These approaches aim for a carefully chosen trade-off between performance and resource consumption in terms of computation and energy.
The development of such approaches is among the major challenges in current machine learning research and key to ensure a smooth transition of machine learning technology from a scientific environment with virtually unlimited computing resources into everyday's applications.
In this article, we provide an overview of the current state of the art of machine learning techniques facilitating these real-world requirements.
In particular, we focus on resource-efficient inference based on deep neural networks (DNNs), the predominant machine learning models of the past decade.
We give a comprehensive overview of the vast literature that can be mainly split into three non-mutually exclusive categories: (i) quantized neural networks, (ii) network pruning, and (iii) structural efficiency.
These techniques can be applied during training or as post-processing, and they are widely used to reduce the computational demands in terms of memory footprint, inference speed, and energy efficiency.
We also briefly discuss different concepts of embedded hardware for DNNs and their compatibility with machine learning techniques as well as potential for energy and latency reduction.
We substantiate our discussion with experiments on well-known benchmark data sets using compression techniques (quantization, pruning) for a set of resource-constrained embedded systems, such as CPUs, GPUs and FPGAs.
The obtained results highlight the difficulty of finding good trade-offs between resource efficiency and prediction quality.
\end{abstract}


\section{Introduction}

Machine learning is a key technology in the 21\textsuperscript{st} century and the main contributing factor for many recent performance boosts in computer vision, natural language processing, speech recognition and signal processing. 
Today, the main application domain and comfort zone of machine learning applications is the ``virtual world'', as found in recommender systems, stock market prediction, and social media services.
However, we are currently witnessing a transition of machine learning moving into ``the wild'', where most prominent examples are autonomous navigation for personal transport and delivery services, and the Internet of Things (IoT).
Evidently, this trend opens several real-world challenges for machine learning engineers.

\begin{figure}[H]
\begin{center}
 \includegraphics[width=11cm]{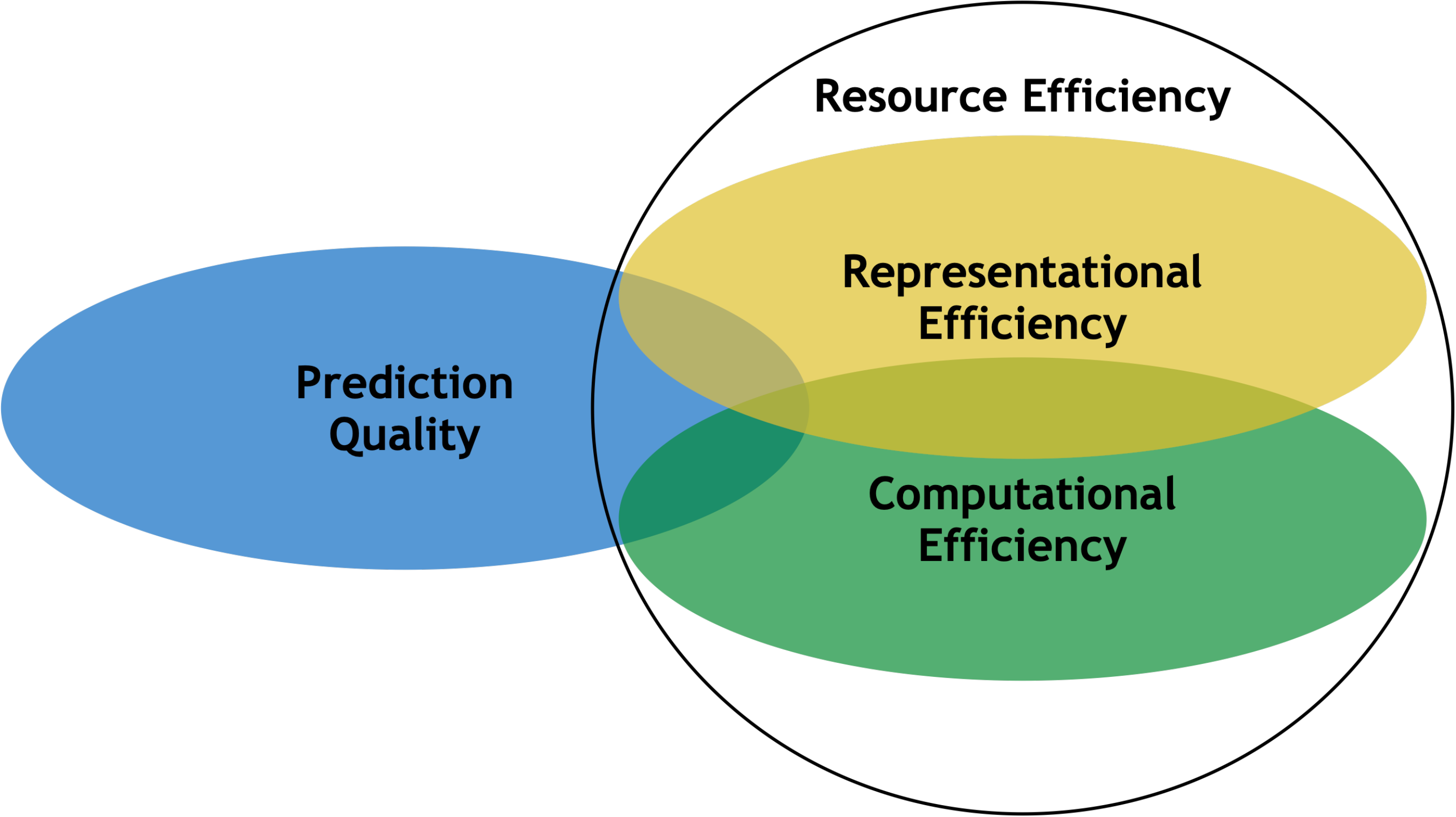}
\caption{Aspects of resource-efficient machine learning models.}
 \label{fig:eff_class}
\end{center}
\end{figure}
Current machine learning approaches prove particularly effective when large amounts of data and ample computing resources are available.
However, in real-world applications the computing infrastructure during the operation phase is typically limited, which effectively rules out most of the current resource-hungry machine learning approaches.
There are several key challenges---illustrated in Figure~\ref{fig:eff_class}---which have to be jointly considered to facilitate machine learning in real-world applications:
\begin{description} 
 \item[Representational efficiency]
 The model complexity in terms of memory footprint should match the (usually limited) resources in deployed systems.
 Model complexity is mainly governed by the employed model and its number of parameters.
 The selected numerical representations and sparsity in the parameters may also have an impact on the memory footprint and, therefore, contribute to the representational efficiency.
 \item[Computational efficiency] The computational cost of performing inference should match the (usually limited) resources in deployed systems and exploit the available hardware optimally in terms of time and energy.
 Computational efficiency, in particular, also includes mapping the representational efficiency to available hardware structures.
 This is in contrast to theoretical inference costs, such as numbers of parameters and required mathematical operations, that often do not reflect inference running time on real hardware well.
 Furthermore, power constraints are key for autonomous and embedded systems, as the device lifetime for a given battery charge needs to be maximized, or constraints set by energy harvesters need to be met.
 \item[Prediction quality]
 The focus of classical machine learning is mostly on optimizing the prediction quality of the models.
 For embedded devices, model complexity versus prediction quality trade-offs must be considered to achieve good prediction performance while simultaneously reducing computational complexity and memory requirements.
\end{description}

In this regard, resource-efficient neural networks for embedded systems are concerned with the trade-off between prediction quality and resource efficiency (i.e., representational efficiency and computational efficiency). This is highlighted in Figure~\ref{fig:eff_class}. 
Note that this requires observing overall constraints such as prediction quality as well as inference latency and/or throughput, chip area and power consumption.

In this article, we review the state of the art in machine learning with regard to these real-world requirements.
We focus on deep neural networks (DNNs), the currently predominant machine learning models.
We formally define DNNs in Section \ref{sec:background} and give a brief introduction to the most prominent building blocks, such as dropout and batch normalization.
While being the driving factor behind many recent success stories, DNNs are notoriously data and resource hungry, a property which has recently renewed significant research interest in resource-efficient approaches.
This paper is dedicated to giving an extensive overview of the current directions of research of these approaches, all of which are concerned with reducing the model size and/or improving inference efficiency while at the same time maintaining accuracy levels close to state-of-the-art models.
We have identified three major directions of research concerned with enhancing resource efficiency in DNNs that we present in Section \ref{sec:literature_overview}.
In particular, these directions are:
\begin{description}
 \item[Quantized Neural Networks]
 Typically, the weights of a DNN are stored as 32-bit float\-ing-point values and during inference millions of floating-point operations are carried out.
 Quantization approaches reduce the number of bits used to store the weights and the activations of DNNs.
 While quantization approaches obviously reduce the memory footprint of a DNN, the selected weight representation potentially also facilitates faster inference using cheaper arithmetic operations.
 Even reducing precision down to binary or ternary values works reasonably well and essentially reduces DNNs to hardware-friendly logical circuits.
 \item[Network Pruning]
 Starting from a fixed, potentially large DNN architecture, pruning approaches remove parts of the architecture during training or after training as a post-processing step.
 The parts being removed range from the very local scale of individual weights---which is called unstructured pruning---to a more global scale of neurons, channels, or even entire layers---which is called structured pruning.
 On the one hand, unstructured pruning is typically less sensitive to accuracy degradation, but special sparse matrix operations are required to obtain a computational benefit.
 On the other hand, structured pruning is more delicate with respect to accuracy but the resulting data structures remain dense such that common highly optimized dense matrix operations, available on most off-the-shelf hardware, can be used.
 \item[Structural Efficiency]
 This category comprises a diverse set of approaches that achieve resource efficiency at the structural level of DNNs.
 \emph{Knowledge distillation} is an approach where a small student DNN is trained to mimic the behavior of a larger teacher DNN, which has been shown to yield improved results compared to training the small DNN directly.
 The idea of \emph{weight sharing} is to use a small set of weights that is shared among several connections of a DNN to reduce the memory footprint.
 Several works have investigated \emph{special matrix structures} that require fewer parameters and allow for faster matrix multiplications---the main workload in fully connected layers.
 Furthermore, there exist several \emph{manually designed architectures} that introduced lightweight building blocks or modified existing building blocks to enhance resource efficiency.
 Most recently, \emph{neural architecture search} methods have emerged that discover efficient DNN architectures automatically.
\end{description}
Evidently, many of the presented techniques are not mutually exclusive and can potentially be combined to further enhance resource efficiency.
For instance, one can both sparsify a model and reduce arithmetic precision.

We complement our literature review with a brief overview of embedded hardware for DNNs in Section~\ref{sec:hardware_overview}.
These hardware platforms can be categorized into CPUs, GPUs, FPGAs and domain-specific accelerators, where each architecture exhibits different properties for deploying models.
We discuss potentials and limitations of such embedded hardware with considerations on vectorization and parallelization, frequency and energy efficiency, as well as their applicability for resource-efficient models.

In Section~\ref{sec:experiments} we substantiate our discussion with experimental results.
We provide a comparison of various quantization approaches for DNNs using the CIFAR-100 data set in Section \ref{sec:quantization_comparison}, followed by an evaluation of prediction quality for different types of pruned structures on the CIFAR-10 data set in Section \ref{sec:pruning}.
We evaluate the inference throughput of the compressed models on an ARM CPU (Section \ref{sec:cpu}), Xilinx FPGA (Section \ref{sec:fpga}) and an embedded NVIDIA GPU (Section \ref{sec:gpu}).
We conclude the experiments with an overall comparison in Section \ref{sec:overall}, where the embedded systems (in combination with compression techniques) are studied with respect to inference throughput and prediction quality.

\section{Background} \label{sec:background}
Before we present a comprehensive overview of the many different techniques for reducing the complexity of DNNs in Section \ref{sec:literature_overview}, this section formally introduces DNNs and some fundamentals required in the remainder of the paper.

\subsection{Feed-forward Deep Neural Networks} \label{sec:background_dnns}
DNNs are typically organized in layers of alternating linear transformations and non-linear activation functions.
A vanilla DNN with $L$ layers is a function mapping an input $\vecx^0$ to an output $y = \vecx^L$ by applying the iterative computation
\begin{align}
 \veca^l &= \matW^l \vecx^{l-1} + \vecb^l, \label{eq:layer_linear} \\
 \vecx^l &= \phi(\veca^l), \label{eq:layer_nonlinear}
\end{align}
where \eqref{eq:layer_linear} computes a linear transformation with weight tensor $\matW^l$ and bias vector $\vecb^l$, and \eqref{eq:layer_nonlinear} computes a non-linear activation function $\phi$ that is typically applied element-wise.
Common choices for $\phi$ are the ReLU function $\phi(a) = \max(a, 0)$, sigmoid functions, such as $\tanh(a)=(e^a - e^{-a})/(e^a + e^{-a})$ and the logistic function $1/(1+e^{-a})$, and, in the context of resource-efficient models, the sign function $\sign(a) = \indicator(a \geq 0) - \indicator(a < 0)$, where $\indicator$ is the indicator function.

In this paper, we focus on hardware-efficient machine learning in the context of classification, i.e., the task of assigning the input $\vecx^0$ to a class $\hat{c} \in \{1, \ldots, \mathcal{C}\}$.
Other predictive tasks, such as regression and multi-label prediction, can be tackled in a similar manner.
For classification tasks, the output activation function $\phi$ for computing $\vecx^L \in \R^{\mathcal{C}}$ is typically the softmax function $\phi(\veca)_i = e^{a_i} / \sum_j e^{a_j}$.
An input $\vecx^0$ is assigned to class $\hat{c} = \argmax_c x^L_c$.

The two most common types of layers are (i) fully connected layers\footnote{Many popular deep learning frameworks refer to fully connected layers as \emph{dense} layers.} and (ii) convolutional layers.
For fully connected layers, the input $\vecx \in \R^{n}$ is a vector whose individual dimensions---also called \emph{neurons}\footnote{For $1<l<L$, we speak of \emph{hidden} layers and \emph{hidden} neurons.}---do not exhibit any a-priori known structure.
The linear transformation of a fully connected layer is implemented as a matrix-vector multiplication $\matW \vecx$ where $\matW \in \R^{m \times n}$.

Convolutions are used if the data exhibits spatial or temporal dimensions such as images, in which case the DNN is called a convolutional neural network (CNN).
Two-dimensional images can be represented as three-dimensional tensors $\vecx^l \in \R^{C \times W \times H}$, where $C$ refers to the number of \emph{channels} (or, equivalently, \emph{feature maps}), and $W$ and $H$ refer to the width and the height of the image, respectively.
A $K_w \times K_h$ convolution using a rank-4 \emph{filter} weight tensor $\matW \in \R^{K_w \times K_h \times C \times D}$ mapping $\vecx^l \in \R^{C \times W \times H}$ to $\veca^{l+1} \in \R^{D \times W \times H}$ is computed as
\begin{align}
 a_{d,w,h}^{l+1} = \sum_{k_w=1}^{K_w} \sum_{k_h=1}^{K_h} \sum_{c=1}^{C} W_{k_w,k_h,c,d} \cdot x^l_{c,i(w,k_w,K_w),i(h,k_h,K_h)}, \label{eq:convolution}
\end{align}
where $i$ is the auxiliary indexing function
\begin{align}
 i(p,k,K) = p - \left\lceil \frac{K}{2} \right\rceil + k.
\end{align}
Each spatial location of the output feature map $\veca^{l+1}$ is computed from a $K_w \times K_h$ region of the input image $\vecx^{l}$.
By using the same filter to compute the values at different spatial locations, a translation invariant detection of features is obtained.
The spatial size of features detected within an image is bounded by the \emph{receptive field}, i.e., the section of the input image that influences the value of a particular spatial location in some hidden layer.
The receptive field is increased by stacking multiple convolutional layers, e.g., performing two consecutive $3 \times 3$ convolutions results in each output spatial location being influenced by a larger $5 \times 5$ region of the input feature maps.

Another form of translational invariance is achieved by \emph{pooling} operations that merge spatially neighboring values within a feature map to reduce the feature map's size.
Common choices are max-pooling and average-pooling which combine the results of neighboring values\footnote{Typically, a $2 \times 2$ region to halve the feature map size is used.} by computing their maximum or average, respectively.
Furthermore, pooling operations also increase the receptive field.

\subsection{Training of Deep Neural Networks} \label{sec:background_training}
The task of training is concerned with adjusting the weights $\matW$ such that the DNN reliably predicts correct classes for unseen inputs $\vecx^0$.
This is accomplished by minimizing a loss function $\loss$ using gradient-based optimization \citep{Nocedal2006}.
Given some labeled training data $\D = \{(\vecx_1^0, t_1), \ldots, (\vecx_N^0, t_N)\}$ containing $N$ input-target pairs, a typical loss function has the form
\begin{align}
 \loss(\matW; \D) = \sum_{n=1}^N l(y(\matW, \vecx_n^0), t_n) + \lambda r(\matW), \label{eq:loss}
\end{align}
where $l(y_n, t_n)$ is the \emph{data term} that penalizes the DNN parameters $\matW$ if the output $y_n$ does not match the target value $t_n$, $r(\matW)$ is a \emph{regularizer} that prevents the DNN from overfitting, and $\lambda > 0$ is a trade-off hyperparameter.
Typical choices for the data term $l(y_n, t_n)$ are the cross-entropy loss or the mean squared error loss, whereas typical choices for the regularizer $r(\matW)$ are the $\ell^1$-norm or the $\ell^2$-norm of the weights.
The loss is minimized using \emph{gradient descent} by iteratively computing
\begin{align}
 \matW \leftarrow \matW - \eta \nabla_{\matW} \loss(\matW; \D), \label{eq:gradient_descent_update}
\end{align}
where $\eta$ is a learning rate hyperparameter.
In practice, more involved \emph{stochastic} gradient descent (SGD) schemes, such as ADAM \citep{Kingma2015b}, are used that randomly select smaller subsets of the data---called mini-batches---to approximate the gradient.

Modern deep learning frameworks play an important role in the growing popularity of DNNs as they make gradient-based optimization particularly convenient:
The user specifies the loss $\loss$ as a computation graph and the gradient $\nabla_{\matW} \loss$ is calculated automatically by the framework using the backpropagation algorithm \citep{Rumelhart1986}.

\subsection{Batch Normalization} \label{sec:background_batchnorm}
The literature has established a consensus that using more layers improves the classification performance of DNNs.
However, increasing the number of layers $L$ also increases the difficulty of training a DNN using gradient-based methods as described in Section \ref{sec:background_training}.
Most modern DNN architecture employ batch normalization \citep{Ioffe2015} after the linear transformation of some or all layers by computing
\begin{align}
 a_{n,d}^l \leftarrow \frac{a_{n,d}^l - \mu_d^l}{\sigma_d^l} \cdot \gamma_d + \beta_d \quad \mbox{with} \quad
 \mu_d^l \leftarrow \frac{1}{N_B} \sum_{n=1}^{N_B} a_{n,d}^l, \quad
 (\sigma_d^l)^2 \leftarrow \frac{1}{N_B - 1} \sum_{n=1}^{N_B} (a_{n,d}^l - \mu_d^l)^2,
\end{align}
where $\beta_d$ and $\gamma_d$ are trainable parameters, and $N_B$ is the mini-batch size of SGD.

The idea is to normalize the activation statistics over the data samples in each layer to zero mean and unit variance.
This results in similar activation statistics throughout the network which facilitates gradient flow during backpropagation.
The linear transformation of the normalized activations with the parameters $\beta$ and $\gamma$ is mainly used to recover the DNNs ability to approximate any desired function---a feature that would be lost if only the normalization step is performed.
Most recent DNN architectures have been shown to benefit from batch normalization, and, as reviewed in Section \ref{sec:literature_structured_pruning}, batch normalization can be targeted to achieve resource efficiency in DNNs.

\subsection{Dropout} \label{sec:background_dropout}
Dropout as introduced by \citet{Srivastava2014} is a way to prevent neural networks from overfitting by injecting multiplicative noise to the inputs of a layer, i.e., $x_d^l \leftarrow x_d^l \cdot \varepsilon$.
A common choice for the injected noise is $\varepsilon \sim \bernoulli(p)$ where the values $x_d^l$ are randomly set to zero with probability $1-p$.
Another prevalent option is Gaussian dropout, where we sample $\varepsilon$ from a normal distribution with mean $1$ and variance $\alpha$.\footnote{The parameters $p$ and $\alpha$ are hyperparameters.}
Intuitively, the idea is that hidden neurons cannot rely on the presence of features computed by other neurons.
Consequently, individual neurons are expected to compute in a sense ``meaningful'' features on their own.
This avoids multiple neurons jointly computing features in an entangled way.
Dropout has been cast into a Bayesian framework which was subsequently exploited to perform network pruning as detailed in Section \ref{sec:literature_bayesian_pruning}.

\subsection{Common Neural Architectures}
\label{sec:background_architectures}

As stated earlier in this section, the majority of architectures follow the simple pattern of repeating several layers of linear transformation followed by a non-linear function $\phi$.
Although most successful architectures follow this scheme, recent architectures have introduced additional components and subtle extensions that have led to new design principles.
In the following, we give a brief overview of the most prominent architectures that have emerged over the past years in chronological order.

\subsubsection{AlexNet}
The AlexNet architecture \citep{Krizhevsky2012} was the first work to show that DNNs are capable of improving performance over conventional hand crafted computer vision techniques by achieving 16.4\% Top-5 error on the ILSVRC12 challenge---an improvement of approximately 10\% absolute error compared to the second best approach in the challenge which relied on well-established computer vision techniques.
This most influential work essentially started the advent of DNNs, which have since spread to virtually all scientific domains, often achieving improvements over well-established methods in their respective fields.

The architecture consists of eight layers---five convolutional layers followed by three fully connected layers.
AlexNet was designed to optimally utilize the available hardware at that time rather than following some clear design principle.
This involves the choice of heterogeneous window sizes $K_w \times K_h$ and seemingly arbitrary numbers of channels per layer $C$.
Furthermore, convolutions are performed in two parallel paths (i.e., grouped convolutions; see Section \ref{sec:literature_architecture_design}) to facilitate the training on two GPUs.

\subsubsection{VGGNet}
The VGGNet architecture \citep{Simonyan14} won the second place at the ILSVRC14 challenge with 7.3\% Top-5 error.
Compared to AlexNet, its structure is more uniform and with up to 19 layers much deeper.
The design of VGGNet is guided by two main principles.
(i) VGGNet uses mostly $3 \times 3$ convolutions and increases the receptive field by stacking several of them.
(ii) After downscaling the spatial dimension with $2 \times 2$ max-pooling, the number of channels should be doubled to avoid information loss.
From a hardware perspective, VGGNet is often preferred over other architectures due to its uniform architecture.

\subsubsection{InceptionNet}
InceptionNet (or, equivalently, GoogLeNet) \citep{Szegedy2015} won the ILSVRC14 challenge with 6.7\% Top-5 error with an even deeper architecture consisting of 22 layers.
The main feature of this architecture is the inception module which combines the outputs of $1 \times 1$, $3 \times 3$, and $5 \times 5$ convolutions by stacking them.
To reduce the computational burden, InceptionNet performs $1 \times 1$ convolutions as proposed in \citep{Lin2014a} to reduce the number of channels immediately before the larger $3 \times 3$ and $5 \times 5$ convolutions.

\subsubsection{ResNet}
Motivated by the observation that adding more layers to very deep conventional CNN architectures does not necessarily reduce the \emph{training} error, residual networks (ResNets) introduced by \citet{He2016} follow a rather different principle.
The key idea is that every layer computes a residual that is \emph{added} to the layer's input.
This is often graphically depicted as a residual path with an attached skip connection.

The authors hypothesize that identity mappings play an important role.
They argue that it is easier to model identity mappings in ResNets by simply setting all the weights of the residual path to zero instead of simulating them by adapting the weights of several consecutive layers in an intertwined way.
In any case, the skip connections reduce the vanishing gradient problem during training and enable extremely deep architectures of up to 152 layers on ImageNet and even up to 1,000 layers on CIFAR-10.
ResNet won the ILSVRC15 challenge with 3.6\% Top-5 error.

\subsubsection{DenseNet}
Inspired by ResNets whose skip connections have shown to reduce the vanishing gradient problem, densely connected CNNs (DenseNets) introduced by \citet{Huang2017} drive this idea even further by connecting each layer to \emph{all} previous layers.
DenseNets are conceptually very similar to ResNets---instead of adding the output of a layer to its input, DenseNets \emph{stack} the output and the input of each layer.
Since this stacking necessarily increases the number of feature maps with each layer, the number of new feature maps computed by each layer is typically small.
Furthermore, it is proposed to use compression layers after downscaling the spatial dimension with pooling, i.e., a $1 \times 1$ convolution is used to reduce the number of feature maps.

Compared to ResNets, DenseNets achieve similar performance, allow for even deeper architectures, and they are more parameter and computation efficient.
However, the DenseNet architecture is highly non-uniform which complicates the hardware mapping and ultimately slows down training.

\subsubsection{MobileNet}
The core idea of MobileNet~\citep{MobileNetv1} is to improve resource efficiency by depthwise-separable convolutions, which decompose an expensive standard convolution into two cheaper sequential convolutions.
In the first convolution, a filter is applied to each channel separately without taking other channels into account.
The second convolution restores information flow across channels by performing a $1 \times 1$ convolution.
By using depthwise-separable convolutions, the number of trainable parameters as well as the number of multiply-accumulate operations (MACs) can be substantially reduced.
It is empirically shown that this has little to no negative impact on prediction quality.
MobileNetV2~\citep{MobileNetv2} extends this concept by introducing additive skip connections and bottleneck layers.
MobileNetV3~\citep{MobileNetv3} extends this even further by also incorporating the neural architecture search (NAS) proposed in MnasNet~\citep{Tan2018}.
More details will be provided in Section~\ref{sec:literature_architecture_design} and Section~\ref{sec:literature_neural_architecture_design}.

\subsubsection{EfficientNet}
EfficientNet~\citep{Tan2019,Tan2021} combines a selection of previous works on resource efficiency with regard to skip connections, depthwise-separable convolutions, Squeeze-and-Excitation (SE) modules, and NAS.

The intuition of Squeeze-and-Excitation networks (SENet)~\citep{Iandola2016} is to learn channel importances which is relevant as the number of channels typically becomes larger with an increasing depth of the architecture.
A squeeze module first reduces information by fusing information from each feature map into a single value, e.g., by global average pooling.
This results in a vector of length $C$, which encodes a feature descriptor of the whole feature map.
This is followed by an excitation module for adaptive recalibration that captures channel-wise dependencies by learning.

NAS is used to find a baseline architecture called EfficientNet-B0.
This model is similar to MnasNet which is composed of multiple stages where each stage is based on one or more layers of depthwise-separable convolutions and SE modules.
While EfficientNet-B0 is the smallest model, enlarged variants of this base model are created by using \emph{compound scaling}.
These upscaled variants are identical to the base model in terms of number of stages and stage type, but for each stage the input feature map resolution, width and depth is scaled.

\subsubsection{Transformers}
While all previously presented architectures are based on convolutional architectures, recently transformers have gained significant attention in natural language processing (NLP), speech processing and computer vision, representing a deviation from traditional sequence processing architectures like recurrent and convolutional neural architectures. 
The core of the transformer architecture~\citep{Transformer} is the self-attention mechanism, which allows to weight the importance of different parts of the input sequence. 
This mechanism enables transformers to capture long-range dependencies in data, which can be challenging for recurrent and convolutional architectures. 
Furthermore, unlike convolutional layers, which have fixed receptive fields, transformers can handle input sequences of variable lengths, making them more flexible and suitable for many tasks.
While transformers were initially developed for sequence-to-sequence tasks in NLP, they have also been explored as replacements for convolutional neural networks in computer vision tasks.

One notable example is the Vision Transformer (ViT)~\citep{VisionTransformer}, where the input image is first divided into non-overlapping patches of fixed size. 
Each patch is then linearly embedded into a flat vector, forming the initial input to the transformer.
To understand the spatial relationships between patches (position of each patch in the image), a positional encoding is added to these embeddings.
The patch embeddings in combination with this positional encoding is then processed by the transformer encoder, such that the self-attention mechanism captures dependencies between different patches and enables the understanding of the global context of the image.
In combination with class tokens, the classification head can then produce classification scores.

While this self-attention mechanism is extremely successful for various tasks, it comes at substantial processing complexity.
A self-attention operation involves computing attention scores for each pair of positions in the input sequence, resulting in a complexity of $N^2$ for a sequence of length $N$.
Similarly, the corresponding attention matrix has a size of $N \times N$.
As transformers typically use multiple attention heads in parallel to capture different aspects of the relationships between input elements, this complexity is further scaled by the number of attention heads $H$.

Sparse attention mechanisms and approximations have been proposed to address this issue and improve the efficiency of transformers for longer sequences.
We refer to the work of \citet{tay2022transformers} which provides an overview of various transformer-based architectures that focus on efficiency, reduced memory-footprint and computational complexity. Most of these methods focus on the quadratic complexity of the self-attention heads and use low-rank matrix operations, downsampling or exploit pre-set or learned sparsity patterns.

\subsection{The Straight-Through Gradient Estimator} \label{sec:background_ste}
Many recently developed methods for resource efficiency in DNNs incorporate components in the computation graph of the loss function $\loss$ that are non-differentiable or whose gradient is zero almost everywhere, such as piecewise constant quantizers.
These components prevent the use of conventional gradient-based optimization as described in Section \ref{sec:background_training}.

\begin{figure}[t]
\begin{center}
 \includegraphics[width=0.7\textwidth]{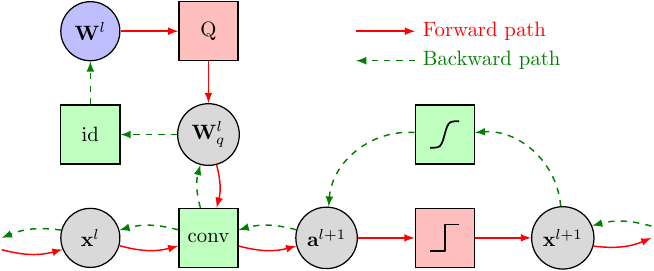}
\caption{A simplified building block of a DNN using the straight-through gradient estimator (STE).
$Q$ denotes some arbitrary piecewise constant quantization function and \emph{id} denotes the identity function which simply passes the gradient on during backpropagation.
In the forward pass, the solid red line is followed which passes the two piecewise constant functions $Q$ and $\sign$ whose gradient is zero almost everywhere (red boxes).
During backpropagation, the dashed green line is followed which avoids these piecewise constant functions and instead only passes differentiable functions (green boxes)---in particular, the functions \emph{id} and $\tanh$ whose shapes are similar to $Q$ and $\sign$ but whose gradient is non-zero.
This allows us to obtain an approximate non-zero gradient for the real-valued parameters $\matW^l$ (blue circle) which are subsequently updated with SGD.
}
 \label{fig:ste}
\end{center}
\end{figure}

The straight-through gradient estimator (STE) is a simple yet effective way to approximate the gradient of such components by simply replacing their gradient with a non-zero value.
Let $f(w)$ be some non-differentiable operation within the computation graph of $\loss$ such that the partial derivative $\partial \loss/\partial w$ is not defined.
The STE then approximates the gradient $\partial \loss/\partial w$ by
\begin{align}
  \frac{\partial \loss}{\partial w} = \frac{\partial \loss}{\partial f} \frac{\partial f}{\partial w} \approx \frac{\partial \loss}{\partial f} \frac{\partial \tilde{f}}{\partial w},
\end{align}
where $\tilde{f}(w)$ is an arbitrary differentiable function with a similar functional shape as $f(w)$.
For instance, in case of the sign activation function $f(w) = \sign(w)$ whose derivative is zero almost everywhere, one could select $\tilde{f}(w) = \tanh(w)$.
Another common choice is the identity function $\tilde{f}(w)=w$ whose derivative is $\tilde{f}^\prime(w) = 1$, which simply passes the gradient on to higher components in the computation graph during backpropagation.
Figure \ref{fig:ste} illustrates the STE applied to a simplified DNN layer.

\subsection{Bayesian Neural Networks} \label{sec:background_bayesian_neural_networks}
Since there exist several works for resource-efficient DNNs that build on the framework of Bayesian neural networks, we briefly introduce the basic principles here.
Given a prior distribution $p(\matW)$ over the weights and a likelihood $p(\D \smallhspace | \smallhspace \matW)$ defined by the softmax output of a DNN as
\begin{align}
 p(\D \smallhspace | \smallhspace \matW) = \prod_{n=1}^N p(y(\matW, \vecx_n^0) = t_n),
\end{align}
we can use Bayes' rule to infer a posterior distribution over the weights, i.e.,
\begin{align}
 p(\matW \smallhspace | \smallhspace \D) = \frac{p(\D \smallhspace | \smallhspace \matW) \ p(\matW)}{p(\D)} \propto p(\D \smallhspace | \smallhspace \matW) \ p(\matW). \label{eq:bayes_rule}
\end{align}
From a Bayesian perspective it is desired to compute expected predictions with respect to the posterior distribution, i.e.,
\begin{align}
  \E_{p(\matW \smallhspace | \smallhspace \D)} [ y(\matW, \vecx^0) ], \label{eq:bayesian_prediction}
\end{align}
and not just to reduce the entire distribution to a single point estimate.
However, due to the highly non-linear nature of DNNs, most exact inference scenarios involving the full posterior $p(\matW \smallhspace | \smallhspace \D)$ are typically intractable and there exist a range of approximation techniques for these tasks, such as variational inference \citep{Hinton1993,Graves2011,Blundell2015} and sampling based approaches \citep{Neal1992}.

Interestingly, training DNNs can often be seen as a very rough Bayesian approximation where we only seek for weights $\matW$ that maximize the posterior $p(\matW \smallhspace | \smallhspace \D)$, which is also known as maximum a-posteriori estimation (MAP).
In particular, in a typical loss $\loss$ as in \eqref{eq:loss} the data term originates from the logarithm of the likelihood $p(\D \smallhspace | \smallhspace \matW)$ whereas the regularizer originates from the logarithm of the prior $p(\matW)$.

A better Bayesian approximation is obtained with variational inference where the aim is to find a variational distribution $q(\matW \smallhspace | \smallhspace \vecnu)$ governed by distribution parameters $\vecnu$ that is as close as possible to the posterior $p(\matW \smallhspace | \smallhspace \D)$ but still simple enough to allow for efficient inference, e.g., for computing $\E_{q(\matW \smallhspace | \smallhspace \vecnu)}[y(\matW, \vecx^0)]$ by sampling from $q$.
This is typically achieved by the so called \emph{mean field} assumption, i.e., by assuming that the weights are independent such that $q(\matW \smallhspace | \smallhspace \vecnu)$ factorizes into a product of factors $q(w \smallhspace | \smallhspace \nu_w)$ for each weight $w \in \matW$.
The most prominent approach to obtain the variational distribution $q(\matW \smallhspace | \smallhspace \vecnu)$ is by minimizing the KL-divergence $\kl(q(\matW \smallhspace | \smallhspace \vecnu)||p(\matW \smallhspace | \smallhspace \D))$ using gradient-based optimization \citep{Ranganath2014,Blundell2015}.

The Bayesian approach is appealing as distributions over the parameters directly translate into predictive distributions.
In contrast to ordinary DNNs that only provide a point estimate prediction, Bayesian neural networks offer predictive uncertainties which are useful to determine how certain the DNN is about its own prediction.
Additionally, the Bayesian framework has got several other useful properties that can be exploited to obtain resource-efficient DNNs.
For instance, the prior $p(\matW)$ allows us to incorporate information about properties, such as sparsity, that we expect to be present in the DNN.
In Section \ref{sec:literature_bayesian_quantization_approaches}, we review weight quantization approaches based on the Bayesian paradigm, and in Section \ref{sec:literature_bayesian_pruning}, we review pruning approaches based on the Bayesian paradigm.

\section{Resource Efficiency in Deep Neural Networks} \label{sec:literature_overview}

In this section, we provide a comprehensive overview of methods that enhance the efficiency of DNNs regarding memory footprint, computation time, and energy requirements.
We have identified three different major approaches that aim to reduce the computational complexity of DNNs, i.e., (i) weight and activation quantization, (ii) network pruning, and (iii) structural efficiency.
These categories are not mutually exclusive, and we present individual methods in the category where their contribution is most significant.

\subsection{Quantized Neural Networks} \label{sec:literature_quantization}
Quantization in DNNs is concerned with reducing the number of bits used for the representation of the weights and the activations.
The reduction in memory requirements are obvious: Using fewer bits for the weights results in a lower memory overhead for storing the corresponding model, and using fewer bits for the activations results in a lower memory overhead for computing predictions.
Furthermore, representations using fewer bits often facilitate faster computation.
For instance, when quantization is driven to the extreme with binary weights $w \in \{-1,1\}$ \emph{and} binary activations $x \in \{-1,1\}$, floating-point or fixed-point dot products are replaced by hardware-friendly logical XNOR and bitcount operations.
In this way, a sophisticated DNN is essentially reduced to a logical circuit.

However, training such discrete-valued DNNs\footnote{Due to finite precision of computer arithmetic, in fact any DNN is discrete-valued. However, we use this term here to emphasize the extremely small number of values.} is difficult as they cannot be directly optimized using gradient-based methods.
The challenge is to reduce the number of bits as much as possible while at the same time keeping the prediction accuracy close to that of a well-tuned full-precision DNN.
Subsequently, we provide a literature overview of approaches that train reduced-precision DNNs, and, in a broader view, we also consider methods that use reduced-precision computations during backpropagation to facilitate low-resource training.

\subsubsection{Early Quantization Approaches} \label{sec:literature_early_quantization}
Approaches for reduced-precision computations date back at least to the early 1990s.
The two works of H{\"o}hfeld and Fahlman \citep{Hoehfeld1992a,Hoehfeld1992b} rounded the weights during training to fixed-point format with different numbers of bits.
They observed that training eventually stalls as small gradient updates are always rounded to zero.
As a remedy, they proposed stochastic rounding, i.e., rounding values to the nearest value with a probability proportional to the distance to the nearest value.
These quantized gradient updates are correct in expectation, do not cause training to stall, and yield good performance with substantially fewer bits than deterministic rounding.
More recently,~\citet{Gupta2015} have shown that stochastic rounding can also be applied to modern deep architectures, as demonstrated on a hardware prototype.

\citet{Lin2015} propose a method to reduce the number of multiplications required during training.
At forward propagation, the weights are stochastically quantized to either binary weights $w \in \{-1,1\}$ or ternary weights $w \in \{-1,0,1\}$ to remove the need for multiplications at all.
During backpropagation, inputs and hidden neurons are quantized to powers of two, reducing multiplications to cheaper bit-shift operations, and leaving only a negligible number of floating-point multiplications to be computed.
However, the speed-up is limited to training since for inference the full-precision weights are required.

\citet{Courbariaux2015a} empirically studied the effect of different numeric formats (i.e., floating-point, fixed-point, and dynamic fixed-point) with varying bit widths on the performance of DNNs.
\citet{Lin2016} consider fixed-point quantization of pre-trained full-precision DNNs.
They formulate a convex optimization problem to minimize the total number of bits required to store the weights and the activations under the constraint that the total output signal-to-quantization noise ratio is larger than a certain prespecified value.
A closed-form solution of the convex objective yields layer-specific bit widths.

\subsubsection{Quantization-aware Training} \label{sec:literature_quantization_aware_training}
Quantization operations, being piecewise constant functions with either undefined or zero derivatives, are not applicable to gradient-based learning using backpropagation.
In recent years, the STE \citep{Bengio2013} (see Section \ref{sec:background_ste}) became the method of choice to compute an approximate gradient for training DNNs with weights that are represented using a very small number of bits.
Such methods typically maintain a set of full-precision weights that are quantized during forward propagation.
During backpropagation, the gradients are propagated through the quantization functions by assuming that their gradient equals one.
In this way, the full-precision weights are updated using gradients computed at the quantized weights.
At test-time, the full-precision weights are abandoned and only the quantized reduced-precision weights are kept.
We term this scheme \emph{quantization-aware training} since quantization is an essential part during forward-propagation and it is intuitive to think of the real-valued weights as becoming robust to quantization.
In a similar manner, many methods employ the STE to approximate the gradient for the quantization of activations.

In \citet{Courbariaux2015b}, binary weight DNNs are trained using the STE to get rid of expensive floating-point multiplications.
They consider deterministic rounding using the sign function and stochastic rounding using probabilities determined by the hard sigmoid function $\max(0,\min(1,(w+1)/2))$.
During backpropagation, a set of auxiliary full-precision weights is updated based on the gradients of the quantized weights.
\citet{Hubara2016} extended this work by also quantizing the activations to a single bit using the sign activation function.
This reduces the computational burden dramatically as floating-point multiplications and additions are reduced to hardware-friendly logical XNOR and bitcount operations, respectively.

\citet{Li2016} trained ternary weights $w \in \{-a,0,a\}$.
Their quantizer sets weights whose magnitude is lower than a certain threshold $\Delta$ to zero, while the remaining weights are set to $-a$ or $a$ according to their sign.
Their approach determines $a>0$ and $\Delta$ \emph{during forward propagation} by approximately minimizing the squared quantization error of the real-valued weights.
\citet{Zhu2017} extended this work to ternary weights $w \in \{-a,0,b\}$ where $a>0$ and $b>0$ are trainable parameters subject to gradient updates.
They propose to select $\Delta^l$ based on the maximum full-precision weight magnitude in each layer, i.e., $\Delta^l = t \cdot \max \{|w| : w \in \matW^l\}$ with $t$ being a hyperparameter.
These asymmetric weights considerably outperform symmetric weights as used by \citet{Li2016}.

\citet{Rastegari2016} approximate full-precision weight filters in CNNs by $\matW = \alpha \matB$ where $\alpha$ is a scalar and $\matB$ is a binary weight matrix.
This reduces the bulk of floating-point multiplications inside the convolutions to either additions or subtractions and only requires a single multiplication per output neuron with the scalar $\alpha$.
In a further step, the layer inputs $\vecx^l$ are quantized in a similar way to perform the convolution using only efficient XNOR and bitcount operations, followed by two floating-point multiplications per output neuron.
Again, the STE is used during backpropagation.
\citet{Lin2017} generalized the ideas of \citet{Rastegari2016} by approximating the full-precision weights using linear combinations of \emph{multiple} binary weight filters for improved classification accuracy.

While most activation binarization methods use the sign function which can be seen as an approximation of the $\tanh$ function, \citet{Cai2017} proposed a half-wave Gaussian quantization that more closely resembles the predominant ReLU activation function.

Motivated by the fact that weights and activations typically exhibit a non-uniform distribution, \citet{Miyashita2016} proposed to quantize values to powers of two.
Their representation allows getting rid of expensive multiplications, and they report higher robustness to quantization than linear rounding schemes using the same number of bits.
\citet{Zhou2017} proposed incremental network quantization where the weights of a pre-trained DNN are first partitioned into two sets.
The weights in the first set are quantized to either zero or powers of two.
The weights in the second set are kept at full precision and retrained to recover from the potential accuracy degradation due to quantization.
They iterate partitioning, quantization, and retraining until all weights are quantized.

\citet{Jacob2018} proposed a quantization scheme that accurately approximates floating-point operations using only integer arithmetic to speed up computation.
During training, their forward pass simulates the quantization step to keep the performance of the quantized DNN close to the performance of using single-precision.
At test-time, weights are represented as 8-bit integer values, reducing the memory footprint by a factor of four.

\citet{Liu2018} introduced Bi-Real net, a ResNet-inspired architecture where the residual path is implemented with efficient binary convolutions while the shortcut path is kept real-valued to preserve the expressiveness of the DNN.
The residual in each layer is computed by first transforming the input with the sign activation, followed by a binary convolution, and a final batch normalization step.

Instead of using a fixed quantizer, in \emph{LQ-Net} \citep{Zhang2018} the quantizer is adapted during training.
The proposed quantizer is inspired by the representation of integers as linear combinations $\vecv^{\transpose} \vecb$ with $\vecv = (2^{0}, \ldots, 2^{K-1})$ and $\vecb \in \{0,1\}^K$.
The key idea is to consider a quantizer that assigns values to the nearest value representable as such a linear combination $\vecv^{\transpose} \vecb$ and to treat $\vecv \in \R^K$ as trainable parameters.
It is shown that such a quantizer is compatible with efficient bit operations.
The quantizer is optimized \emph{during forward propagation} by minimizing the quantization error objective $\|\matB \vecv - \vecx\|^2$ for $\matB \in \{-1,1\}^{N \times K}$ and $\vecv$ by alternately fixing $\matB$ and minimizing $\vecv$ and vice versa.
It is proposed to use layer-wise quantizers for the activations and channel-wise quantizers for the weights, i.e., an individual quantizer for each layer and channel, respectively.

Relaxed Quantization \citep{Louizos2019} introduces a stochastic differentiable soft rounding scheme.
By injecting additive noise to the deterministic weights before rounding, one can compute probabilities of the weights being rounded to specific values in a predefined discrete set.
Subsequently, these probabilities are used to differentiably round the weights using the Gumbel-softmax approximation \citep{Jang2017}.
Since this soft rounding scheme produces only values that are close to values from the discrete set but which are not exactly from this set, the authors also propose a hard variant using the STE.

\citet{Dong2019} introduced Hessian-aware mixed-precision quantization for DNNs.
Their method quantifies the sensitivity of individual DNN blocks to weight quantization using the largest eigenvalue of the block-wise Hessian matrices which can be computed using the power iteration method.
They compute two different orderings of the individual DNN blocks, both of which are based on these eigenvalues.
The first ordering determines a relative ordering of the bit widths of individual blocks.
This substantially reduces the exponential search space of layer-specific weights and allows them to manually set appropriate bit widths.
The second ordering takes these bit widths into account and determines the sequence in which blocks are quantized and fine-tuned using quantization-aware training.

In general, a linear quantizer has three characteristic properties, (i) a step size $Q_d$, (ii) a dynamic range $Q_{\mathrm{max}}$, and (iii) the number of bits $Q_b$.
Since these quantities are interrelated according to
\begin{align}
Q_{\mathrm{max}} = (2^{Q_b-1} - 1) \smallhspace Q_d, \label{eq:linear_quantizer_relation}
\end{align}
it is specified by knowing any two of them \citep{Uhlich2020}.
Given fixed layer-wise bit widths $Q_b^l$, \citet{Esser2020} incorporated layerwise step sizes $Q_d^l$ as trainable parameters in the computation graph.
By training the step sizes $Q_d^l$ using the STE, they are adapted to the given objective.
This is in contrast to previous work, such as XNOR-Net \citep{Rastegari2016}, that determine the step size $Q_d^l$ using certain statistics obtained from the values to be quantized.
\citet{Uhlich2020} extended this idea to mixed-precision quantization.
They investigated the three different possibilities to specify a linear quantizer \eqref{eq:linear_quantizer_relation} by only two of its characteristic properties and discovered substantial differences in the training behavior.
They propose to parameterize the quantizers using the step size $Q_d^l$ and the dynamic range $Q_{\mathrm{max}}^l$ and to train these values using backpropagation and the STE to obtain layerwise bit widths $Q_b^l$.

There also exist works that perform quantization \emph{during} backpropagation to facilitate resource-efficient training.
\citet{Zhou2016} presented several quantization schemes for the weights and the activations that allow for flexible bit widths.
Furthermore, they also propose a quantization scheme for backpropagation to facilitate low-resource training.
In accordance with earlier work mentioned above, they note that stochastic quantization is essential for their approach.
In \citet{Wu2018}, weights, activations, weight gradients, and activation gradients are subject to customized quantization schemes that allow for variable bit widths and facilitate integer arithmetic during training and testing.
In contrast to \citet{Zhou2016}, the work of \citet{Wu2018} accumulates weight changes to low-precision weights instead of full-precision weights.

While most work on quantization based approaches is empirical, some works gained more theoretical insights \citep{Li2017,Anderson2018}.
The recent work of \citet{Shekhovtsov2020} has shown that for stochastic binary networks the STE arises from particular linearization approximations.

Even for large language models~\citep{touvron2023llama,DBLP:journals/corr/abs-2005-14165,DBLP:journals/corr/abs-1909-08053} quantization can offer benefits.
However, for such large models quantization-aware retraining is impractical.
\citet{frantar2023optq} proposed a one-shot post-training quantization scheme for transformers which supports low bit-width quantization and can compress transformer-based language models within few GPU hours.
Targeting the same problem, \citet{Lin2023AWQ} introduced activation-aware weight quantization, which exploits the fact that the weights of large language models are not equally important.
They propose to guide the selection of important weights by activation magnitudes (rather than weight magnitudes) and protecting salient weights by per-channel scaling.

\subsubsection{Bayesian Approaches for Quantization} \label{sec:literature_bayesian_quantization_approaches}
In this section, we review some quantization approaches, most of which are closely related to the Bayesian variational inference framework (see Section \ref{sec:background_bayesian_neural_networks}).

The work of \citet{Achterhold2018} builds on the variational dropout based pruning approach of \citet{Louizos2017} (see Section \ref{sec:literature_bayesian_pruning}).
They introduce a mixture of log-uniform priors whose mixtures are centered at predefined quantization values.
Consequently, the approximate posterior also concentrates at these values such that weights can be safely quantized without requiring a fine-tuning procedure.

The following works in this section directly operate on \emph{discrete} weight distributions and, consequently, do not require a rounding procedure.
\citet{Soudry2014} approximate the true posterior $p(\matW \smallhspace | \smallhspace \D)$ over discrete weights using expectation propagation \citep{Minka2001} with closed-form online updates.
Starting with an uninformative approximation $q(\matW \smallhspace | \smallhspace \vecnu)$, their approach combines the current approximation $q(\matW \smallhspace | \smallhspace \vecnu)$ (serving as the prior in Bayes' rule \eqref{eq:bayes_rule}) with the likelihood for a single-sample data set $\D_n = \{(\vecx_n^0,t_n)\}$ to obtain a refined posterior.
By proposing several approximations, they obtain a closed-form refinement step.

Although deviating from the Bayesian variational inference framework as no similarity measure to the true posterior is optimized, the approach of \citet{Shayer2018} trains a distribution $q(\matW \smallhspace | \smallhspace \vecnu)$ over either binary weights $w \in \{-1,1\}$ or ternary weights $w \in \{-1,0,1\}$.
They propose to minimize an expected loss $\E_{q(\matW \smallhspace | \smallhspace \vecnu)}[\loss(\matW; \D)]$ for the variational parameters $\vecnu$ with gradient-based optimization using the local reparameterization trick \citep{Kingma2015a}.
After training has finished, the discrete weights are obtained by either sampling or taking a mode from $q(\matW \smallhspace | \smallhspace \vecnu)$.
Since their approach is limited to the ReLU activation function, \citet{Peters2018} extended their work to the $\sign$ activation function.
This involves several non-trivial changes since the sign activation, due to its zero derivative, requires that the local reparameterization trick must be performed \emph{after} the $\sign$ function.
Consequently, \emph{distributions} need to be propagated through commonly used building blocks such as batch normalization and pooling operations.
\citet{Roth2019} further extended these works to beyond three distinct discrete weights, and they introduced some technical improvements.

\Citet{vanBaalen2020} propose a Bayesian mixed-precision quantization method for power-of-two bit widths.
Their method is based on a recursive view of quantization where residual quantization errors are repeatedly quantized.
They introduce gates that determine how many recursive quantization steps should be performed which in turn determines the number of used bits.
While the quantization itself is subject to the STE, they propose to train gate probabilities using the Bayesian variational inference framework.
The use of fewer bits for quantization is encouraged using a specific prior and, through an additional zero-bit gate, their framework simultaneously allows for weight pruning.

\citet{Havasi2019} introduced a novel Bayesian compression technique that we present here in this section although it is rather a coding technique than a quantization technique.
In a nutshell, their approach first computes a variational distribution $q(\matW \smallhspace | \smallhspace \vecnu)$ over real-valued weights using mean field variational inference and then it encodes a sample from $q(\matW \smallhspace | \smallhspace \vecnu)$ in a smart way.
They construct an approximation $\tilde{q}(\matW)$ to $q(\matW \smallhspace | \smallhspace \vecnu)$ by importance sampling using the prior $p(\matW)$ as
\begin{align}
  q(\matW \smallhspace | \smallhspace \vecnu) \approx \tilde{q}(\matW) = \sum_{i=1}^{2^K} \frac{q(\matW_i \smallhspace | \smallhspace \vecnu)}{p(\matW_i)} \delta_{\matW_i}(\matW) \qquad \mbox{with} \qquad \matW_i \sim p(\matW), \label{eq:miracle_importance_sampling}
\end{align}
where $\delta_{\matW_i}$ denotes a point mass located at $\matW_i$.
In the next step, a sample $\matW$ from $\tilde{q}(\matW)$ (or, equivalently, an approximate sample from $q(\matW \smallhspace | \smallhspace \vecnu)$) is drawn which can be encoded by the corresponding number $k \in \{1,\ldots,2^K\}$ using $K$ bits.
Using the same random number generator initialized with the same seed as in \eqref{eq:miracle_importance_sampling}, the weights $\matW$ can be recovered by sampling $2^K$ weights $\matW_i$ from the prior $p(\matW)$ and selecting $\matW_k$.
Since the number of samples $2^K$ required to obtain a reasonable approximation to $q(\matW \smallhspace | \smallhspace \vecnu)$ in \eqref{eq:miracle_importance_sampling} grows exponentially with the number of weights, this sampling based compression scheme is performed for smaller weight blocks such that each weight block can be encoded with $K$ bits.

\subsection{Network Pruning} \label{sec:literature_network_pruning}
Network pruning methods aim to achieve parameter sparsity by setting a substantial number of DNN weights to zero.
Subsequently, the sparsity is exploited to improve resource efficiency of the DNN.
On the one hand, there exist \emph{unstructured} pruning approaches that set individual weights, regardless of their location in a weight tensor, to zero.
Unstructured pruning approaches are typically less sensitive to accuracy degradation, but they require special sparse tensor data structures that in turn yield practical efficiency improvements only for very high sparsity.
On the other hand, \emph{structured} pruning methods aim to set whole weight structures to zero, e.g., by setting \emph{all} weights of a matrix column to zero we would effectively prune an entire neuron.
Conceptually, structured pruning is equivalent to removing tensor dimensions such that the reduced tensor remains compatible with highly optimized dense tensor operations.

In this section, we start with the unstructured case which includes many of the earlier approaches and continue with structured pruning that has been the focus of more recent works.
Then we review approaches that relate to Bayesian principles before we discuss approaches that prune structures dynamically during forward propagation.

\subsubsection{Unstructured Pruning} \label{sec:literature_unstructured_pruning}
One of the earliest approaches to reduce the network size is the \emph{optimal brain damage} algorithm of \citet{LeCun1989}.
Their main finding is that pruning based on weight magnitude is suboptimal, and they propose a pruning scheme based on the increase in loss function.
Assuming a pre-trained network, a local second-order Taylor expansion with a diagonal Hessian approximation is employed that allows us to estimate the change in loss function caused by weight pruning without re-evaluating the costly network function.
Removing parameters is alternated with retraining the pruned network.
In this way, the model size can be reduced substantially without deteriorating its performance.
\citet{Hassibi1992} found the diagonal Hessian approximation to be too restrictive, and their \emph{optimal brain surgeon} algorithm uses an approximated full covariance matrix instead.
While their method, similar as in \citet{LeCun1989}, prunes weights that cause the least increase in loss function, the remaining weights are simultaneously adapted to compensate for the negative effect of weight pruning.
This bypasses the need to alternate several times between pruning and retraining the pruned network.

However, it is not clear whether these approaches scale up to modern DNN architectures since computing the required (diagonal) Hessians is substantially more demanding (if not intractable) for millions of weights.
Therefore, many of the more recently proposed techniques still resort to magnitude-based pruning.
\citet{Han2015} alternate between pruning connections below a certain magnitude threshold and re-training the pruned DNN.
The results of this simple strategy are impressive, as the number of parameters in pruned DNNs is an order of magnitude smaller (9$\times$ for AlexNet and $13\times$ for VGG-16) than in the original networks.
Hence, this work shows that DNNs are often heavily over-parameterized.
In a follow-up paper, \citet{Han2016} proposed \emph{deep compression}, which extends the work \citet{Han2015} by a parameter quantization and parameter sharing step, followed by Huffman coding to exploit the non-uniform weight distribution.
This approach yields a reduction in memory footprint by a factor of 35--49 and, consequently, a reduction in energy consumption by a factor of 3--5.

\citet{Guo2016} discovered that irreversible pruning decisions limit the achievable sparsity and that it is useful to reincorporate weights pruned in an earlier stage.
In addition to each dense weight matrix $\matW \in \R^{m \times n}$, they maintain a corresponding binary mask matrix $\matT \in \{0,1\}^{m \times n}$ that determines whether a weight is currently pruned or not.
In particular, the actual weights used during forward propagation are obtained as $\matW \odot \matT$ where $\odot$ denotes element-wise multiplication.
Their method alternates between updating the weights $\matW$ based on gradient descent, and updating the weight masks $\matT$ by thresholding the real-valued weights according to
\begin{align}
 T_{i,j}^{t+1} = \begin{cases} 0 \qquad & \mbox{if } |W_{i,j}^t| \in [0,a) \\ T_{i,j}^{t} \qquad &\mbox{if } |W_{i,j}^t| \in [a,b) \\ 1 \qquad & \mbox{if } |W_{i,j}^t| \in [b,\infty)  \end{cases}, \label{eq:guo2016_treshold}
\end{align}
where $a$ and $b$ are thresholds and $t$ refers to the iteration number.
Most importantly, weight updates are also applied to the currently pruned weights according to $\matT$ using the STE, such that pruned weights can reappear in \eqref{eq:guo2016_treshold}.
This reduces the number of parameters of AlexNet by a factor of 17.7 without deteriorating performance.

\citet{Gadhikar2023randompruning} provide theoretical insights and have demonstrated empirically that sparsity can also be introduced by using a fixed but randomly selected sparsity pattern.
While random pruning is computationally inexpensive, it may not achieve optimal sparsity.
However, it still holds potential for further refinement, making it valuable as initial sparse random masks can bypass the computationally expensive process of pruning a dense network from scratch.

\subsubsection{Structured Pruning} \label{sec:literature_structured_pruning}
In \citet{Mariet2016}, a determinantal point process (DPP) is used to find a group of neurons that are diverse and exhibit little redundancy.
Conceptually, a DPP for a given ground set $\mathcal{S}$ defines a distribution over subsets $S \subseteq \mathcal{S}$ where subsets containing diverse elements have high probability.
They consider $\mathcal{S}$ to be the set of $N$-dimensional vectors that individual neurons compute over the whole data set.
Their approach samples a diverse set of neurons $S \subseteq \mathcal{S}$ according to the DPP and then prunes the other neurons $\mathcal{S} \setminus S$.
To compensate for the negative effect of pruning, the outgoing weights of the remaining neurons after pruning are adapted so as to minimize the activation change of the next layer.

\citet{Wen2016} incorporated group lasso regularizers in the objective to obtain different kinds of sparsity in the course of training.
They were able to remove filters, channels, and even entire layers in architectures containing skip connections.
\citet{Liu2017} proposed to introduce an $\ell^1$-norm regularizer on the scale parameters $\gamma$ of batch normalization and to set $\gamma=0$ by thresholding.
Since each batch normalization parameter $\gamma$ corresponds to a particular channel in the network, this results in channel pruning with minimal changes to existing training pipelines.
In \citet{Huang2018}, the outputs of different structures are scaled with individual trainable scaling factors.
By using a sparsity enforcing $\ell^1$-norm regularizer on these scaling factors, the outputs of the corresponding structures are driven to zero and can be pruned.

Rather than pruning based on small parameter values, ThiNet \citep{Luo2017} is a data-driven approach that prunes channels having the least impact on the subsequent layer.
To prune channels in layer $l$, they propose to sample several activations $x_{d,w,h}^{l+1}$ at randomly selected spatial locations $(w,h)$ and channels $d$ of the following layer, and to greedily prune channels whose removal results in the least increase of squared error over these randomly selected activations.
After pruning, they adapt the remaining filters to minimize the squared reconstruction error by minimizing a least squares problem.

\citet{Louizos2018} propose to multiply weights with stochastic binary 0-1 gates associated with trainable probability parameters that effectively determine whether a weight should be pruned or not.
They formulate an expected loss with respect to the distribution over the stochastic binary gates.
By incorporating an expected $\ell^0$-norm regularizer over the weights, the probability parameters associated with these gates are encouraged to be close to zero.
To enable the use of the reparameterization trick, a continuous relaxation of the binary gates using a modified binary Gumbel-softmax distribution is used \citep{Jang2017}.
They show that their approach can be used for structured sparsity by associating the stochastic gates to entire structures such as channels.
\citet{Li2019} extended this work by using the recently proposed unbiased ARM gradient estimator \citep{Yin2019} instead of using the biased Gumbel-softmax approximation.

While typically a large, over-parameterized network is subject to one or multiple iterations of training, pruning, and re-training, \citet{Liu2019} argue that for many neural architectures, training on the fixed sparse architecture from scratch yields good results.
Their experiments imply that often the sparse structure itself is important, rather than the specific weights that remain after pruning.
Furthermore, this highlights a connection between pruning and neural architecture search.

\subsubsection{Bayesian Pruning} \label{sec:literature_bayesian_pruning}
In \citet{Graves2011} and \citet{Blundell2015}, mean field variational inference is employed to obtain a factorized Gaussian approximation $q(\matW \smallhspace | \smallhspace \vecnu)$, i.e., instead of learning a deterministic weight $w$ per connection, they train for each connection a weight mean $w_{\mu}$ and a weight variance $w_{\sigma^2}$.
After training, weights are pruned by thresholding the ``signal-to-noise ratio''~$|w_{\mu}/w_{\sigma}|$.

Some pruning approaches are based on variational dropout \citep{Kingma2015a} which interprets dropout as performing variational inference with specific prior and approximate posterior distributions.
Within this framework, the otherwise fixed dropout rates $\alpha$ of Gaussian dropout appear as free parameters that can be optimized to improve a variational lower bound.
\citet{Molchanov2017} exploited this freedom to optimize individual weight dropout rates $w_\alpha$ such that weights $w$ can be safely pruned if their dropout rate $w_\alpha$ is large.
This idea has been extended by \citet{Louizos2017} by using sparsity enforcing priors and assigning dropout rates to groups of weights that are all connected to the same structure, which in turn allows for structured pruning.
Furthermore, they show how their approach can be used to determine an appropriate bit width for each weight by exploiting the well-known connection between Bayesian inference and the minimum description length (MDL) principle \citep{Gruenwald2007}.

\subsubsection{Dynamic Network Pruning} \label{sec:literature_dynamic_pruning}
So far, we have presented methods that result in a fixed reduced architecture.
In the following, we present methods that determine dynamically in the course of forward propagation which structures should be computed or, equivalently, which structures should be pruned.
The intuition behind this idea is to vary the time spent for computing predictions based on the difficulty of the given input samples $\vecx^0$.

\citet{Lin2017b} proposed to train, in addition to the DNN, a recurrent neural network (RNN) as decision network which determines the channels to be computed using reinforcement learning.
To keep the decision network lightweight and fast, it is only fed with compressed feature maps using global pooling.
Thus, the RNN can aggregate state information over the layers to compute its pruning decisions, saving enough computations to compensate for the overhead and thus being more efficient than static alternatives.

In \citet{Dong2017}, convolutional layers of a DNN are extended by a parallel low-cost convolution whose output after the ReLU function is used to scale the outputs of the potentially high-cost convolution.
Due to the ReLU function, several outputs of the low-cost convolution will be exactly zero such that the computation of the corresponding output of the high-cost convolution can be omitted.
For the low-cost convolution, they propose to use weight tensors $\matW \in \R^{1 \times 1 \times C \times D}$ and $\matW \in \R^{K \times K \times C \times 1}$.
However, practical speed-ups are only reported for the $K \times K$ convolution where all channels at a given spatial location might get set to zero.

In a similar approach proposed by \citet{Gao2019}, the spatial dimensions of a feature map are reduced by global average pooling to a vector $\veca \in \R^C$ which is linearly transformed to $\vecb \in \R^{D}$ using a single low-cost fully connected layer.
To obtain a sparse vector $\vecc \in \R^{D}$, $\vecb$ is fed into the ReLU function, followed by a $k$-winner-takes-all function that sets all entries of a vector to zero that are not among the $k$ largest entries in absolute value.
By multiplying $\vecc$ in a channel-wise manner to the output of a high-cost convolution, at least $D-k$ channels will be zero and need not be computed.
The number of channels $k$ is derived from a predefined minimal pruning ratio hyperparameter.

\subsection{Structural Efficiency in DNNs} \label{sec:literature_structural_efficiency}
In this section, we review strategies that establish certain structural properties in DNNs to improve computational efficiency.
Each of the proposed subcategories in this section follows rather different principles and the individual techniques might not be mutually exclusive.

\subsubsection{Weight Sharing} \label{sec:literature_weight_sharing}
Another technique to reduce the model size is weight sharing.
We note that weight sharing and quantization methods (see Section \ref{sec:literature_quantization}) are closely related:
Quantization methods often have an inherent weight sharing property since the number of possible quantization values is often much smaller than the number of weights.
However, the purpose of a method is typically different depending on which category it belongs to.
On the one hand, the focus of weight quantization methods typically lies on the employed numerical formats.
The purpose of these formats is to reduce the storage per weight and to facilitate more efficient computations.
Furthermore, the number of distinct weight values is typically rather small and fixed, and the particular weight values are often constrained or even fixed in advance.
On the other hand, the purpose of weight sharing is to reduce the memory by reducing the overall number of distinct weight values.
For these methods, the particular weight values typically remain unconstrained.
Note that some methods cannot be clearly attributed to either category, e.g., in deep compression \citep{Han2016} weight sharing and quantization are in part used synonymously.

\citet{Chen2015} used a hashing function to randomly group network connections into ``buckets'', where the connections in each bucket share the same weight value.
The advantage of their approach is that weight assignments need not be stored explicitly since they are given implicitly by the hashing function.
The authors show a memory footprint reduction by a factor of 10 while keeping the prediction quality essentially unaffected. 

\citet{Ullrich2017} extended the soft weight sharing approach proposed by \citet{Nowlan1992} to achieve both weight sharing and sparsity.
The idea is to select a Gaussian mixture model prior over the weights and to train both the weights as well as the parameters of the mixture components.
During training, the mixture components collapse to point measures and each weight gets attracted by a certain weight component.
After training, weight sharing is obtained by assigning each weight to the mean of the component that best explains it, and weight pruning is obtained by assigning a relatively high mixture mass to a component with a fixed mean at zero.

\citet{Roth2018b} utilized weight sharing to reduce the memory footprint of a large Bayesian ensemble of DNNs.
The weight sharing is enforced by introducing a Dirichlet process prior over the weight prior distribution.
They propose a sampling based inference scheme by alternately sampling weight assignments using Gibbs sampling and sampling weights using Hamiltonian Monte Carlo \citep{Neal1992}.
By using the same weight assignments for multiple weight samples, the memory overhead for the weight assignments becomes negligible and the total memory footprint of an ensemble is reduced.

\subsubsection{Knowledge Distillation} \label{sec:literature_knowledge_distillation}
Knowledge distillation is a method where the knowledge contained in a large \emph{teacher} model is transferred to a smaller \emph{student} model.
In the first step, a large teacher model is obtained with conventional training methods on the given training data.
Subsequently, the smaller student model is trained on data where the ground truth labels have been replaced by the soft labels obtained from the output of the teacher model, e.g., from the softmax output of a DNN.
It has been shown that this substantially increases the accuracy of the student model compared to directly training on the given training data.

This general scheme is model agnostic, and early works applied knowledge distillation to compress \emph{ensembles} of shallow neural networks \citep{Zeng2000} and other types of classifiers \citep{Bucila2006} into a single neural network.
\citet{Zeng2000} have shown that training on soft labels obtained from the teacher results in higher accuracy than training on the actual hard predictions.
The work of \citet{Bucila2006} emphasizes the ability to train the student on unlabeled data to further reduce the accuracy gap between student and teacher.
In addition, they presented a method to generate new synthetic inputs from the given training set, which might be useful if additional unlabeled data is limited or not available.
They showed that the accuracy of the student can improve substantially when trained on these synthetically generated inputs.

\citet{Ba2014} applied these ideas to investigate the importance of depth in a DNN.
They trained shallower (but not necessarily smaller) neural networks by mimicking the output activations $\veca^L$ produced by a teacher DNN before applying the softmax function.
The resulting shallow models perform similar to their deeper counterparts which was not achievable by training the shallow model on the ground truth targets directly.
Therefore, the authors conclude that shallower models are as expressive as deeper models but they are more difficult to train.

The work of \citet{Li2014} and \citet{Hinton2015} applied knowledge distillation with the main focus on reducing model complexity of a large teacher DNN.
\citet{Hinton2015} proposed to obtain the soft labels $\hat{\vecy}$ from the teacher by scaling the output activations with a temperature $\tau > 0$ as
\begin{align}
  \hat{y}_i = \frac{\exp(a_i^L / \tau)}{\sum_j \exp(a_j^L / \tau)}. \label{eq:knowledge_distillation_softened_softlabels}
\end{align}
For $\tau > 1$, the labels tend to become more uniform which has been reported to facilitate training.
Furthermore, they propose to utilize the ground truth labels by minimizing a weighted average of the traditional cross-entropy loss based on the ground truth labels $t$ and the knowledge distillation loss based on the soft targets $\hat{\vecy}$ in \eqref{eq:knowledge_distillation_softened_softlabels}.
Notably, it was the work of \citet{Hinton2015} that coined the term \emph{knowledge distillation}.

FitNets \citep{Romero2015} extend these ideas by also transferring knowledge from intermediate layers.
They select an intermediate layer from the teacher DNN as the \emph{hint layer} which they try to mimic in an intermediate \emph{guide layer} of the student DNN.
Since the hint layer and the guide layer are generally of different size, they introduce a regressor that predicts the hint layer from the guide layer.
This ensures that the guide layer contains the same information as the hint layer.
The proposed procedure operates in two stages.
In the first stage, the student is trained up to the guide layer by minimizing the discrepancy between them.
In the second stage, the whole student DNN is trained using conventional knowledge distillation as in \citet{Hinton2015}.

\citet{Kim2018} argue that matching the raw features of certain intermediate layers as in \citet{Romero2015} is suboptimal since it is difficult to compare individual layers of different DNNs.
Therefore, they propose a method to match more understandable \emph{factors} extracted from the intermediate layers of the student and the teacher DNNs.
Starting from a pre-trained teacher DNN, they first train an autoencoder which they call \emph{paraphraser} to extract understandable factors from a selected intermediate layer of the teacher DNN.
The student DNN is extended by a regressor which they call \emph{translator} whose purpose is to predict the paraphraser factors from the features of a selected intermediate layer.
The student DNN is then trained to simultaneously minimize the cross-entropy loss on the ground truth labels and the difference between paraphraser and translator output.
They employ the paraphraser and the translator after the last convolutional layer in their DNNs.

In the context of quantization, knowledge distillation has been used to reduce the accuracy gap between real-valued DNNs and quantized DNNs \citep{Mishra2018,Polino2018}.
In particular, a real-valued teacher DNN is used to improve the accuracy of a quantized student DNN.
\citet{Mishra2018} showed improved results using three different modes of knowledge distillation training, including a mode where the student and the teacher are trained simultaneously from scratch.

\citet{Phuong2019} transferred knowledge between different parts of the \emph{same} model.
They employ multi-exit architectures which provide anytime predictions after certain intermediate layers; therefore, allowing for a trade-off between accuracy and prediction latency at run-time.
The knowledge from the (most accurate) final layer is transferred to the earlier exits to improve their accuracy.
Furthermore, they show that the earlier layers can be trained with unlabeled data in a semi-supervised setting.

\citet{Kaliamoorthi2021pQRNN} demonstrated the effectiveness of knowledge distillation for large language models.
They proposed a new efficient student architecture consisting of projection, bottleneck, convolution and pooling layers. This projection-based embedding-free architecture is lightweight in terms of latency and parameter footprint.  

In a Bayesian context, \citet{Korattikara2015} applied knowledge distillation to condense a large ensemble of DNNs, for instance, obtained by sampling from the posterior distribution $p(\matW \smallhspace | \smallhspace \D)$.
In this way, expected predictions \eqref{eq:bayesian_prediction} obtained by averaging the outputs of the individual models can be transferred to a single DNN.
Their method trains a single student DNN using the outputs of teacher DNNs that are generated on the fly using SGLD \citep{Welling2011}.

\subsubsection{Special Matrix Structures} \label{sec:literature_special_matrix_structures}
In this section, we review approaches that aim to reduce the model size by employing efficient matrix representations.
There exist several methods using low-rank decompositions which represent a large matrix (or a large tensor) using only a fraction of the parameters.
In most cases, the implicitly represented matrix is never computed explicitly such that also a computational speed-up is achieved.
Furthermore, there exist approaches using special matrices that are specified by only few parameters and whose structure allows for extremely efficient matrix multiplications.

\citet{Denil2013} proposed a method that is motivated by training only a subset of the weights and predicting the values of the other weights from this subset.
In particular, they represent weight matrices $\matW \in \R^{m \times n}$ using a low-rank approximation $\matU \matV$ with $\matU \in \R^{m \times k}$, $\matV \in \R^{k \times n}$, and $k < \min\{m,n\}$ to reduce the number of parameters.
Instead of learning both factors $\matU$ and $\matV$, prior knowledge, such as smoothness of pixel intensities in an image, is incorporated to compute a fixed $\matV$ using kernel-techniques or auto-encoders, and only the factor $\matU$ is learned.

In \citet{Novikov2015}, the tensor train matrix format is employed to substantially reduce the number of parameters required to represent large weight matrices of fully connected layers.
Their approach enables the training of very large fully connected layers with relatively few parameters, and they achieve improved performance compared to simple low-rank approximations.

\citet{Denton2014} propose specific low-rank approximations and clustering techniques for individual layers of pre-trained CNNs to reduce both memory footprint and computational overhead.
Their approach yields substantial improvements for both the computational bottleneck in the convolutional layers and the memory bottleneck in the fully connected layers.
By fine-tuning after applying their approximations, the performance degradation is kept at a decent level.
\citet{Jaderberg2014} propose two different methods to approximate pre-trained CNN filters as combinations of rank-1 basis filters to speed up computation.
The rank-1 basis filters are obtained either by minimizing a reconstruction error of the original filters or by minimizing a reconstruction error of the outputs of the convolutional layers.
\citet{Lebedev2015} approximate the convolution tensor using the canonical polyadic (CP) decomposition---a generalization of low-rank matrix decompositions to tensors---using non-linear least squares.
Subsequently, the convolution using this low-rank approximation is performed by four consecutive convolutions, each with a smaller filter, to reduce the computation time substantially.

In \citet{Cheng2015}, the weight matrices of fully connected layers are restricted to circulant matrices $\matW \in \R^{n \times n}$, which are fully specified by only $n$ parameters.
While this dramatically reduces the memory footprint of fully connected layers, circulant matrices also facilitate faster computation as matrix-vector multiplication can be efficiently computed using the fast Fourier transform.
In a similar vein, \citet{Yang2015} reparameterize matrices $\matW \in \R^{n \times n}$ of fully connected layers using the Fastfood transform as $\matW{=}\mathbf{SHG \Pi HB}$, where $\mathbf{S}$, $\mathbf{G}$, and $\mathbf{B}$ are diagonal matrices, $\mathbf{\Pi}$ is a random permutation matrix, and $\mathbf{H}$ is the Walsh-Hadamard matrix.
This reparameterization requires only a total of $4 n$ parameters, and similar as in \citet{Cheng2015}, the fast Hadamard transform enables an efficient computation of matrix-vector products.

\subsubsection{Manual Architecture Design} \label{sec:literature_architecture_design}
Instead of modifying existing architectures to make them more efficient, manual architecture design is concerned with the development of new architectures that are inherently resource-efficient.
Over the past years, several design principles and building blocks for DNN architectures have emerged that exhibit favorable computational properties and sometimes also improve performance.

CNN architectures are typically designed to have a transition from convolutional layers to fully connected layers.
At this transition, activations at all spatial locations of each channel are typically used as individual input features for the following fully connected layer.
Since the number of these features is typically large, there is a memory bottleneck for storing the parameters of the weight matrix especially in the first fully connected layer.

\citet{Lin2014a} introduced two concepts that have been widely adopted by subsequent works.
The first technique, \emph{global average pooling}, largely solves the above-mentioned memory issue at the transition to fully connected layers.
Global average pooling reduces the spatial dimensions of each channel into a single feature by averaging over all values within a channel.
This reduces the number of features at the transition drastically, and by having the same number of channels as there are classes, it can also be used to completely remove fully connected layers.
Secondly, they used $1 \times 1$ convolutions with weight kernels $\matW \in \R^{1 \times 1 \times C \times D}$ which can be seen as performing the operation of a fully connected layer over each spatial location across all channels.

These $1 \times 1$ convolutions have been adopted by several popular architectures \citep{Szegedy2015,He2016,Huang2017} and, due to their favorable computational properties compared to convolutions that take a spatial neighborhood into account, later have also been exploited to improve computational efficiency.
For instance, InceptionNet \citep{Szegedy2015} proposed to split standard $K \times K$ convolutions into two cheaper convolutions: (i) a $1 \times 1$ convolution to reduce the number of channels such that (ii) a subsequent $K \times K$ convolution is performed faster.
Similar ideas are used in \emph{SqueezeNet} \citep{Iandola2016} which employs $1 \times 1$ convolutions to reduce the number of input channels of subsequent parallel $1 \times 1$ and $3 \times 3$ convolutions.
In addition, SqueezeNet uses the global average pooling output of per-class channels directly as input to the softmax in order to avoid fully connected layers that typically consume the most memory.
On top of that, they also applied deep compression \citep{Han2016} (see Section \ref{sec:literature_unstructured_pruning}) to reduce the memory footprint of their model even further.

\citet{Szegedy2016} extended the InceptionNet architecture by spatially separable convolutions to reduce the computational complexity, i.e., a $K \times K$ convolution is split into a $K \times 1$ convolution followed by a $1 \times K$ convolution.
In \emph{MobileNet} \citep{Howard2017} depthwise separable convolutions are used to split a standard convolution in another way: (i) a depthwise convolution and (ii) a $1 \times 1$ convolution.
The depthwise convolution applies a $K \times K$ filter to each channel separately without taking the other channels into account whereas the $1 \times 1$ convolution then aggregates information across channels.
Although these two cheaper convolutions together are less expressive than a standard convolution, they can be used to trade off a small loss in prediction accuracy with a drastic reduction in computational overhead and memory requirements.

\citet{Sandler2018} extended these ideas in their \emph{MobileNetV2} to an architecture with residual connections.
A typical residual block with bottleneck structure in ResNet \citep{He2016} contains a $1 \times 1$ bottleneck convolution to reduce the number of channels, followed by a $3 \times 3$ convolution, followed by another $1 \times 1$ convolution to restore the original number of channels again.
Contrary to that building block, MobileNetV2 introduces an \emph{inverted} bottleneck structure where the shortcut path contains the bottleneck and the residual path performs computations in a high-dimensional space.
In particular, the residual path performs a $1 \times 1$ convolution to \emph{increase} the number of channels, followed by a cheap \emph{depthwise} $3 \times 3$ convolution, followed by another $1 \times 1$ convolution to reduce the number of channels again.
They show that their inverted structure is more memory efficient since the shortcut path, which needs to be kept in memory during computation of the residual path, is considerably smaller.
Furthermore, they show improved performance compared to the standard bottleneck structure.

While it was more of a technical detail rather than a contribution on its own, AlexNet \citep{Krizhevsky2012} used \emph{grouped convolutions} with two groups to facilitate model parallelism for training on two GPUs with relatively low memory capacity.
Instead of computing a convolution using a weight tensor $\matW \in \R^{K \times K \times gC \times gD}$, a grouped convolution splits the input into $g$ groups of $C$ channels that are independently processed using weight tensors $\matW_g \in \R^{K \times K \times C \times D}$.
The outputs of these $g$ convolutions are then stacked again such that the same number of input and output channels are maintained while considerably reducing the computational overhead and memory footprint.

Although this reduces the expressiveness of the convolutional layer since there is no interaction between the different groups, \citet{Xie2017} used grouped convolutions to enlarge the number of channels of a ResNet model which resulted in accuracy gains while keeping the computational complexity of the original ResNet model approximately the same.
\citet{Zhang2018b} introduced a ResNet-inspired architecture called \emph{ShuffleNet} which employs $1 \times 1$ \emph{grouped} convolutions since $1 \times 1$ convolutions have been identified as computational bottlenecks in previous works, e.g., see \citet{Howard2017}.
To combine the computational efficiency of grouped convolutions with the expressiveness of a full convolution, ShuffleNet incorporates \emph{channel shuffle} operations after grouped convolutions to partly recover the interaction between different groups.

Over-parameterization is a typical concern for many vision tasks, and architectural efficiency and compression support progress in learning.
This is in contrast to most natural language tasks where the trend is towards ever larger models.
For large language models like PaLM~\citep{chowdhery2022palm} where 6144 TPUs are used for training, efficient scheduling and communication between hardware resources is essential.

\subsubsection{Neural Architecture Search} 
\label{sec:literature_neural_architecture_design}

Neural architecture search (NAS) is a recently emerging field concerned with the automatic discovery of good DNN architectures.
This is achieved by designing a discrete space of possible architectures in which we subsequently search for an architecture that optimizes some objective---typically the validation error.
By incorporating a measure of resource efficiency into this objective, this technique has recently attracted attention for the automatic discovery of resource-efficient architectures.

The task is very challenging:
On the one hand, evaluating the validation error is time-consuming as it requires a full training run and typically only results in a noisy estimate thereof.
On the other hand, the space of architectures is typically of exponential size in the number of layers.
Hence, the space of architectures needs to be carefully designed in order to facilitate an efficient search within it.

The influential work of \citet{Zoph2017} introduced a scheme to encode DNN architectures of arbitrary depth as sequences of tokens which can be sampled from a controller RNN.
This controller RNN is trained with reinforcement learning to generate well performing architectures using the validation error on a held-out validation set as a reward signal.
However, the training effort is enormous since more than 10,000 training runs are required to achieve state-of-the-art performance on CIFAR-10.
This would be impractical on larger data sets such as ImageNet which was partly solved by subsequent NAS approaches, e.g., in \citet{Zoph2018}.
In this review, we highlight methods that also consider resource efficiency constraints for NAS.

In MnasNet \citep{Tan2018}, a RNN controller is trained by also considering the latency of the sampled DNN architecture measured on a real mobile device.
They achieve performance improvements under predefined latency constraints on a specific device.
To run MnasNet on the large-scale data sets ImageNet and COCO \citep{Lin2014b}, their algorithm is run on a \emph{proxy task} by only training for five epochs, and only the most promising DNN architectures were trained using more epochs.

\citet{Wang2019} determined the individual bit widths of mixed-precision quantization using a similar reinforcement learning framework.
Their controller DNN generates for each layer two bit widths, one for the weights and one for the activations.
A pre-trained full-precision DNN is then quantized using these bit widths and fine-tuned for one epoch to obtain a reward signal that is subsequently used to update the controller.
Their method incorporates hardware-specific constraints, such as latency and energy consumption, that must be met by the controller.

Instead of generating architectures using a controller, ProxylessNAS \citep{Cai2019} uses a heavily over-parameterized model where each layer contains several parallel paths, each computing a different architectural block with its individual parameters.
For each layer, probability parameters for selecting a particular architectural block are introduced which are trained via backpropagation using the STE.
After training, the most probable path determines the selected architecture.
To favor resource-efficient architectures, a latency model is build using measurements done on a specific real device whose predicted latencies are used as a differentiable regularizer in the cost function.
In their experiments, they show that different target devices prefer individual DNN architectures to obtain a low latency.

Instead of using a different path for different operations in each layer, single-path NAS \citep{Stamoulis2019} combines all operations in a single \emph{shared weight superblock} such that each operation uses a subset of this superblock.
A weight-magnitude-based decision using trainable threshold parameters determines which operation should be performed, allowing for gradient-based training of both the weight parameters and the architecture.
Again, the STE is employed to backpropagate through the threshold function.

\citet{Wu2018b} performed mixed-precision quantization using similar NAS concepts to those used by \citet{Liu2019b} and \citet{Cai2019}.
They introduce gates for every layer that determine the number of bits used for quantization, and they perform continuous stochastic optimization of probability parameters associated with each of these gates.

\citet{Liu2019} have replicated several experiments of pruning approaches (see Section \ref{sec:literature_network_pruning}) and they observed that the typical workflow of training, pruning, and fine-tuning is often not necessary and only the discovered sparsity structure is important.
In particular, they show for several pruning approaches that randomly initializing the weights after pruning and training the pruned structure from scratch results in most cases in a similar performance as performing fine-tuning after pruning.
They conclude that network pruning can also be seen as a paradigm for architecture search.

\citet{Tan2019} proposed EfficientNet which employs NAS for finding a resource-efficient architecture as a key component.
In the first step, they perform NAS to discover a small resource-efficient model which is much cheaper than searching for a large model directly.
In the next step, the discovered model is enlarged by a principled compound scaling approach which simultaneously increases the number of layers, the number of channels, and the spatial resolution.
Although this approach is not targeting resource efficiency on its own, EfficientNet achieves state-of-the-art performance on ImageNet using a relatively small model.

\citet{Galen} proposed Galen, a NAS framework to search effective layer-specific compression parameters for quantization and pruning by considering experimentation on hardware targets.
Thereby selected parameters are tested for the task at hand on hardware architectures, and feedback in terms of inference latency is used in the optimization objective.
Moreover, a sensitivity analysis is used as proxy to estimate the impact of compression on prediction accuracy.
This enables an automated compression of models tailored to given hardware targets balancing multiple compression methods.

\citet{elsken2019neural} provides a detailed overview of neural architectures search methods highlighting the key dimensions of search space, search strategy and strategies to estimate performance.

\section{Embedded Hardware for Deep Neural Networks} \label{sec:hardware_overview}
Improvements in hardware for deep learning are a key driver for the recent success stories of AI applications through DNNs.
Both training and inference have extremely high demands on their targeted platform and certain hardware requirements can be the deciding factor whether an application can be realized.
This section briefly introduces the most important hardware for deep learning and discusses their potentials and limitations.
While this discussion is generic and independent from training or inference, it should be noted that all processor concepts are available in different scales, ranging from mobile to server variants.

\subsection{CPUs}
CPUs were originally designed to optimize single-thread performance in order to execute an individual computation within the shortest possible latency.
Unfortunately, single-thread performance is stagnating since the end of Dennard scaling \citep{dennard}, and now performance scaling usually requires parallelization.
While multithreading is a rather obvious solution for parallelization that is applicable to many tasks, 
vectorization is a technique that promises great potential for certain applications.
Vectorization applies a single instruction to multiple pre-selected data elements and, thus, avoids costly at-runtime dependency checking while maximizing instruction reuse.
CPUs show excellent properties of exploiting sparse DNNs due to their short vector units and the low amount of multithreading together with high frequency.
Furthermore, they usually support 8-bit integer formats and feature certain instructions for extremely low representation and, consequently, they are well suited for quantization operations.

\subsection{GPUs}
GPUs were initially designed to accelerate image and video processing only and are nowadays the most popular general-purpose accelerators for many tasks, such as scientific and AI computations.
The architecture consists of many streaming multiprocessors which are highly parallel and each implements many lightweight cores.
Thus, GPUs are massively parallel processors with large memory that provides extremely high bandwidth and throughput, but significantly lower frequency in comparison to CPUs.
The extremely high amount of parallelism and the resulting demand on structured computations, however, virtually prevents the deployment of sparse computations.
Modern GPU designs and their respective software stack implement support for reduced-precision computations, such as 8-bit integer and half-precision floating-point formats, which are very well suited for deep learning.
More extreme forms of quantization are not yet supported and do not result in more efficient inference or training.

\subsection{FPGAs}
Field-programmable gate arrays (FPGAs) are a family of processors that implement a large array of configurable logic blocks which can be programmed using hardware description languages (e.g., VHDL, Verilog, HLS).
This concept is the main difference to ASICs in terms of technology since the hardware can be designed for specific functional or application requirements.
While this reconfigurability enables various opportunities that go beyond the capabilities of CPUs and GPUs, it comes at the cost of much lower frequency and reduced on-chip memory.
FPGAs are in principle very well suited for DNNs, since compute units can be specifically tailored to fit the diverse computations while also enabling massive amounts of parallelism.
Reconfigurable hardware is especially interesting for compressed DNNs due to their flexibility to implement any data format as well as sparse logic.

\subsection{Domain-Specific Accelerators}
Recent interest in deep learning has motivated to push advancements in the development of custom accelerators, such as Google's TPUs and Graphcore's IPU.
The key feature of the TPU (and most of the other deep learning accelerators) is a 256$\times$256 matrix-multiplication unit that is referred to as \emph{systolic array}.
Systolic arrays are a variant of massively parallel processor arrays that are very suitable for regular tasks, such as linear algebra operations, and a promising candidate to address the increasing costs of data movements.
The objective of such arrays is to minimize instruction fetch and data access costs by constraining the data flow to matrix and vector operations.
However, data movements can only be reduced if locality effects are sufficiently exploited and the data flow constraints of a systolic array may result in poor utilization and latency increase.
Such domain-specific accelerators are usually highly constrained when aiming to optimize DNNs through compression.
For instance, the TPU supports 8-bit integer and half-precision floating-point formats while other (potentially lower-precision) representations are not efficiently supported by hardware.
Furthermore, the dense structure of the systolic arrays demands for similarly dense computations and cannot exploit fine-grained sparsity patterns.

\subsection{Loop-Back vs.\ Data-Flow Architectures}
One can roughly categorize hardware platforms for deep learning inference into loop-back and data-flow architectures.
Loop-back architectures use a fixed processor and memory system to move data from off-chip memory to the processor and leverage the available compute resources. 
This is performed for each layer or operation sequentially until inference computation has finished. 
The drawback of loop-back architectures is that they potentially require many data movements from and to off-chip memory, which is time and energy consuming.
CPUs aim to reduce these memory accesses by featuring large on-chip caches and reuse data as much as possible.
Similar are domain-specific accelerators, such as TPUs, which usually feature a large and programmable scratch pad memory on chip.
On the contrary, GPUs feature large register files and aim to hide memory latency by leveraging parallel slackness.
Another critical aspect of loop-back architectures is low compute utilization, which can potentially occur if certain layer or operation types do not fit the static compute array (i.e., if operation size is too low).
The advantage of such a \emph{generic} compute architecture is that they allow arbitrary operations in combination with productive code generation since the hardware does not need to be optimized for a certain task.
Continuous improvements in semi-conductor and processor technology are the main improvement factor of such inference engines.

In contrast to this, data-flow architectures use a reconfigurable processor and memory system for computing the inference.
Here, each layer or operation within a neural architecture is assigned a dedicated compute engine and its own memory subsystem in order to enable inference in a pipelined fashion.
This avoids off-chip accesses for intermediate operations completely by simple forwarding the computed results to the next hardware layer.
Furthermore, data-flow architectures achieve excellent utilization of the available hardware logic, since several compute engines can be tailored to the required operation type and latency.
One drawback of this data-flow architecture is, however, that it requires long development costs because it does not only require software but also hardware optimizations.
In addition, reconfigurable hardware comes at the cost of reduced absolute compute power in comparison to ASIC designs.
The main limitation of data-flow architectures is that they demand the entire neural architecture (weights and activations) to be stored on chip, which is highly restrictive for large models.

\section{Experimental Results}   \label{sec:experiments}
We provide experimental results for modern DNN architectures trained on well-known benchmark data sets.
The focus of our experiments is on quantization (see Section \ref{sec:literature_quantization}) and structured pruning approaches (see Section \ref{sec:literature_structured_pruning}) since they are among the earliest and most efficient approaches to enhance the computational efficiency of DNNs.

We compare several quantization approaches discussed in this paper in terms of prediction quality in Section \ref{sec:quantization_comparison}.
Next, we compare different DNN architectures and pruning structures (i.e., the type of structure considered for pruning, such as channels) using model metrics such as number of computations and memory requirements in Section~\ref{sec:pruning}.
Furthermore, we benchmark the compressed models on mobile variants of CPU (Section~\ref{sec:cpu}), FPGA (Section~\ref{sec:fpga}), and GPU (Section~\ref{sec:gpu}), and provide an overall comparison in Section~\ref{sec:overall}.

While domain-specific accelerators, such as Google's TPU, excel in their specific performance, they are usually limited to a set of specific operations and are neither flexible in terms of data types nor sparse calculations. Furthermore, in particular for the TPU, experimentation is often hindered due to limitations in the tool chain which is not flexible enough to support such optimizations. They are not suited to execute generic compressed models and are therefore not included in the following experiments.

Similarly, there are tools such as Tensorflow Lite (TFLite) which is a library to deploy neural architectures on resource-constrained devices such as mobile or embedded devices.
Being a library severely limits the possibility to implement different optimization techniques with regard to resource efficiency.
We therefore chose to not cover TFLite in this work.
Instead, the interested reader can find an excellent overview of the practical use of TFLite for mobile phones (as well as another review on resource efficiency) in \citet{Menghani2023survey}.

The main focus of this section is to showcase the difficulty of finding good trade-offs between prediction quality and resource efficiency, which is formed by the combination of representational efficiency and computational efficiency.
As this paper is mainly dedicated to giving a comprehensive literature overview of the current state of the art, an extensive evaluation of the many presented methods in Section \ref{sec:literature_overview} would be infeasible and it is also not within the scope of this paper.

\subsection{Prediction Quality of Compressed DNNs}
\label{sec:cifar}

\subsubsection{Prediction Quality using Different Quantization Approaches}   
\label{sec:quantization_comparison}
In the first experiment we compare the performance of several quantization approaches.
We use a DenseNet architecture \citep{Huang2017} consisting of 100 layers with bottleneck and compression layers, i.e., a DenseNet-BC-100.
We select the default growth rate of $k=12$ for the model, i.e., the number of feature maps added per layer.
We conduct our experiments on the CIFAR-100 data set where the task is to classify RGB images of size 32$\times$32 pixels to one of 100 object categories.
The CIFAR-100 data set is split into 50,000 training images and 10,000 test images.
We selected some of the most popular quantization approaches (see Section \ref{sec:literature_quantization}) for our comparison: binary weight networks (BWN) \citep{Courbariaux2015b}, binarized neural networks (BNN) \citep{Hubara2016}, DoReFa-Net \citep{Zhou2016}, trained ternary quantization (TTQ) \citep{Zhu2017}, and LQ-Net~\citep{Zhang2018}.
For this experiment, we quantize the DNNs in three different modes: (i) weight-only, (ii) activation-only, and (iii) combined weight and activation quantization.
However, note that some quantization approaches are designed for a particular mode, e.g., BWN and TTQ only consider weight quantization whereas BNN only considers combined weight and activation quantization.

\begin{figure}[t]
  \begin{center}
    \includegraphics[width=0.7\textwidth]{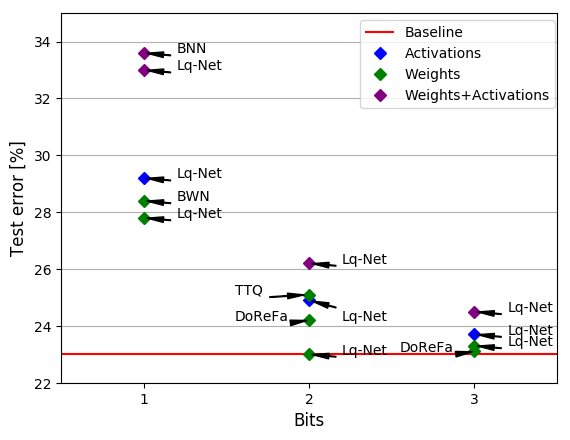}
    \caption{Comparison of several popular quantization approaches (see Section \ref{sec:literature_quantization}) using the DenseNet-BC-100 architecture trained on the CIFAR-100 data set.
    The horizontal red line shows the error of the real-valued baseline.
    Quantization is performed using different bit widths in three different modes: activation-only (blue), weight-only (green), and combined weight and activation quantization (purple).}
    \label{fig:comparison}
  \end{center}
\end{figure}

Figure~\ref{fig:comparison} reports the test errors for different bit widths of the selected quantization approaches.
The horizontal red line shows the test error of the real-valued baseline DenseNet-BC-100.
For combined weight and activation quantization we use the same bit widths for the weights and the activations.

As expected, the test error decreases gradually with increasing bit widths for all quantization modes and for all quantization approaches.
Furthermore, the results indicate that prediction performance is more sensitive to activation quantization than to weight quantization, which is in line with the results reported by many works reviewed in Section \ref{sec:literature_quantization}.

The more advanced LQ-Net approach clearly outperforms the rather simple linear quantization of DoReFa-Net and the specialized binary and ternary approaches.
However, this performance improvement comes at the cost of longer training times.
For instance, the training time per iteration increases---in relation to training without quantization---for DoReFa-Net by a factor of 1.5 compared to a factor of up to 4.6 (depending on the bit width) for LQ-Net.

\subsubsection{Prediction Quality using Different Pruning Structures}
\label{sec:pruning}
In the next experiment, we explore the performance metrics of different DNN architectures (ResNet and DenseNet) and pruning structures (such as channels, kernels and groups) on the CIFAR-10 task.
CIFAR-10 is similar to CIFAR-100 used in the previous section (i.e., image size and size of training and test sets are equal) except that it contains only ten object classes.
We use wide residual networks (WRNs) by \citet{WRN} with a depth of 28 layers, one of the best performing architectures on this task.
This architecture is identical to the original ResNet model except that it is scaled in width rather than depth.
Additionally, we create a DenseNet variant for this experiment which is scaled in depth to 28 layers and the width is varied until it approximately matches the number of parameters and computations of the WRN model in order to guarantee a fair comparison.
We apply parameterized structured pruning (PSP) by \citet{LOD20}, a method that allows to dynamically learn the shape of DNNs through structured sparsity.
PSP parameterizes arbitrary structures in a weight tensor and leverages weight decay to detect unimportant structures that can be pruned.
In this experiment, we select pruning structures that are in line with commonly used DNN libraries for convolutions: we use channel pruning to learn the number of input and output feature maps, kernel pruning to learn the size of the convolution kernel, and group pruning to learn heterogeneous group sizes for grouped convolutions (see Section \ref{sec:literature_architecture_design} for a discussion on grouped convolutions).

\begin{figure*}
	\centering
	\begin{subfigure}{.45\textwidth}
		\centering
		\includegraphics[width=1.0\linewidth]{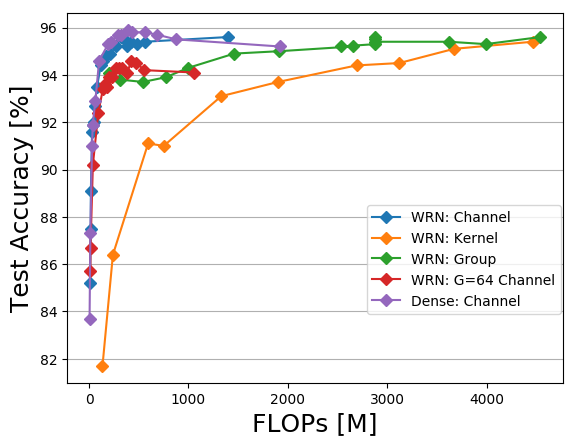}
		\label{fig:as_cifar0}
	\end{subfigure}
	\begin{subfigure}{.45\textwidth}
		\centering
		\includegraphics[width=1.0\linewidth]{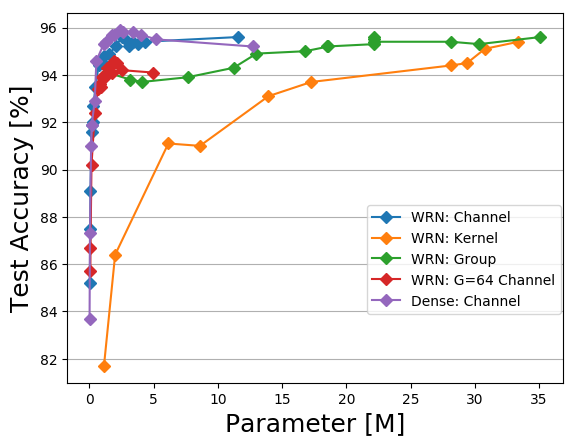}
		\label{fig:as_cifar1}
	\end{subfigure}
	
	\begin{subfigure}{.45\textwidth}
		\centering
		\includegraphics[width=1.0\linewidth]{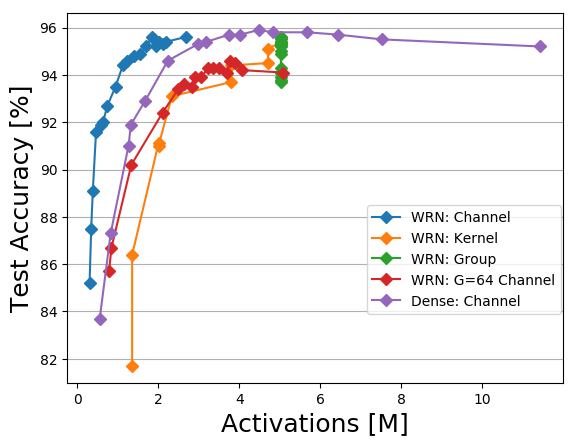}
		\label{fig:as_cifar2}
	\end{subfigure}
	\begin{subfigure}{.45\textwidth}
		\centering
		\includegraphics[width=1.0\linewidth]{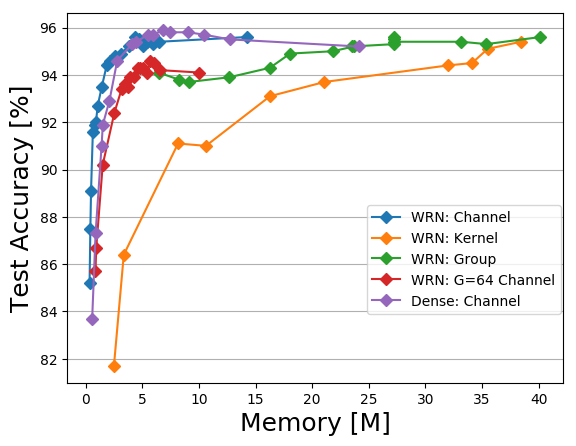}
		\label{fig:as_cifar3}
	\end{subfigure}
	\caption{Performance metrics of several pruning structures and DNN architectures on CIFAR-10 for WRNs and DenseNets.}
	\label{fig:as_cifar}
\end{figure*}

In a first step, channel (\emph{WRN: Channel}), kernel (\emph{WRN: Kernel}), and group pruning (\emph{WRN: Group}) are evaluated separately on the WRN architecture.
The results for number of floating-point operations (FLOPs), parameters, activations, and memory ($=$ parameters $+$ activations) are reported in Figure~\ref{fig:as_cifar}.
When considering the number of FLOPs and parameters, which are the main metrics in the literature on resource-efficient DNNs, it is clear that channel and group pruning significantly outperform kernel pruning.
This indicates a high sensitivity of the kernel size to the overall accuracy. 
Group and channel pruning perform very similar with respect to FLOPs and parameters, especially for highly compressed models. 
However, when also considering the number of activations and overall memory consumption (i.e., parameters and activations), group pruning performs significantly worse since grouped convolutions only remove connections from input channels to output channels while keeping the overall number of channels the same.
As a result, channel pruning is the best performing compression structure when applied in isolation.

In a next step, the group size is set to 64 and channel pruning is applied (\emph{WRN: G=64 Channel}), in order to evaluate the performance when different sparse structures are combined.
This combination performs worse than pure channel pruning for all metrics and requires a large amount of activations. 
Ultimately, it can be stated that group convolutions are excellent at reducing FLOPs and parameters but can harm the overall memory requirements by increasing the amount of activations.

Lastly, the DenseNet variant is compressed using channel pruning (\emph{Dense: Channel}).
The dense architecture outperforms the residual blocks in terms of number of FLOPs as well as parameters.
In terms of number of activations, however, residual blocks are clearly more beneficial, which also influences the overall memory consumption. 
In summary, one can observe that DenseNets are more parameter/computation-efficient and ResNets are more memory-efficient.

\subsection{Evaluating Compressed DNNs on Embedded Hardware}

\subsubsection{Evaluating Compressed DNNs on CPU}
\label{sec:cpu}
\begin{figure}[t]
	\begin{center}
		\includegraphics[width=0.7\textwidth]{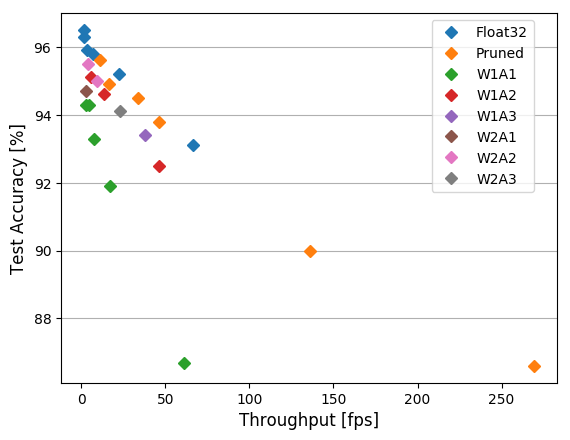}
		\caption{Throughput-accuracy trade-off of pruned and quantized WRN models on the CIFAR-10 task for an ARM CPU.}
		\label{Tradeoff}
	\end{center}
\end{figure}

Section~\ref{sec:cifar} explored the impact of several network quantization approaches and structured pruning on the prediction quality.
In this section. we use the well-performing LQ-Net approach for quantization and PSP (for channel pruning) to measure the inference throughput of the quantized and pruned models separately on an ARM Cortex-A53 CPU.
The WRN model on the CIFAR-10 task is used again as a baseline, with a depth of 28 layers, varying widths of the model, and weights/activations quantized to different bit widths.
Figure~\ref{Tradeoff} reports test accuracies and throughput for different WRN variants and compression methods.
Please note that multiple points for the same bit width correspond to a different width scaling of the model.

The results reveal that quantization does not provide throughput improvements on this processor.
This is mainly due to the efficient floating-point units within the CPU in combination with fast on-chip memory and the high overhead resulting from performing low-bit-width computations.
Moreover, the dimensions of some layers are too small to fit well to the bit level vectorized instructions and these layers limit the overall performance.
Together with the accuracy reduction, low-bit-width quantization does not yield convincing results on this processor.
Quantized DNNs with 1-bit weights and activations are the worst performing models, which is due to the severe implications of extreme quantization on prediction performance.
As can be seen, however, the overall performance of the quantized models increases considerably when the bit width of activations is increased to 2--3 bit whilst the bit width of the weights is kept low. 
On the contrary, channel pruning consistently performs equally to the baseline model with respect to accuracy and throughput.
Pruning is therefore the more suitable compression technique for this embedded processor, especially when considering that CPUs could potentially leverage a much finer sparsity granularity.

\subsubsection{Evaluating Quantized DNNs on FPGAs}
\label{sec:fpga}
\begin{figure}[t]
	\begin{center}
		\includegraphics[width=0.7\textwidth]{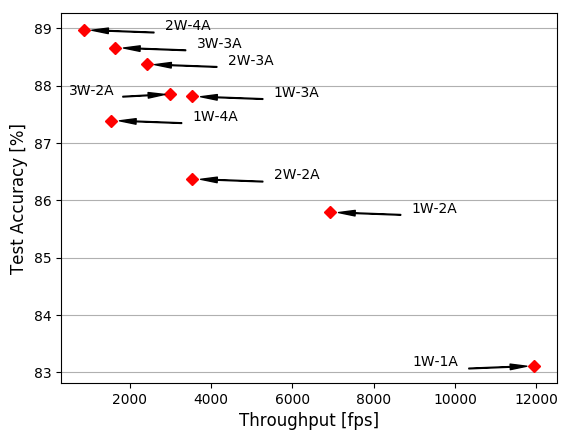}
          \caption{Throughput-accuracy trade-off of different quantized VGG models on the CIFAR-10 task for an FPGA data-flow architecture.}
		\label{fig:finn}
	\end{center}
\end{figure}

While the previous section indicates that quantized DNNs do not provide throughput improvements on general-purpose processors without explicit hardware support, there are other hardware platforms where quantization is mandatory.
Data-flow architectures, as found typically on FPGAs, where the main objective is to keep all required data for inference in on-chip memory, are usually constrained by the requirements for weight as well as activation storage.
This section evaluates quantized DNNs on FPGAs using the FINN framework \citep{Umuroglu2017} for generating data-flow architectures on reconfigurable hardware. 
Figure~\ref{fig:finn} shows test accuracy over throughput of the FINN data-flow architectures mapped to a XILINX Ultra96 FPGA using different bit combinations.
A variant of the VGG architecture is used on the CIFAR-10 task for evaluation because FINN does not support residual connections yet, and the configuration of the FINN framework is adjusted so that highest throughput is targeted with respect to the available resources of the device~(BRAM, LUTs, etc). 

As expected, the test accuracy increases gradually with high bit widths while the throughput decreases accordingly.
Following the Pareto front starting from the bottom right indicates that the best performing models use a combination of 1 bit for the weights and a gradual increase of activations up to 3 bits.
Afterwards the models perform best if the weights are scaled to 2 bits and the activation bit width is further increased to 4 bits. 
This supports the observation of the previous sections, showing that model accuracy is sensitive to activation quantization rather than weight quantization.

\subsubsection{Evaluating Pruned DNNs on GPU}
\label{sec:gpu}
\begin{figure}[t]
	\begin{center}
		\includegraphics[width=0.7\textwidth]{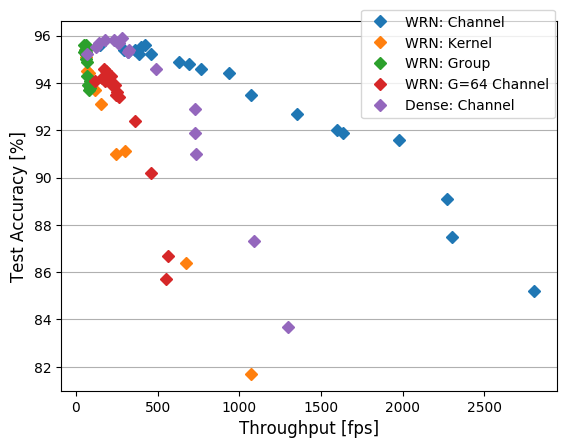}
		\caption{Throughput-accuracy trade-off of different pruning methods for an embedded GPU on the CIFAR-10 task.}
		\label{fig:throughput_cifar}
	\end{center}
\end{figure}

Section~\ref{sec:pruning} explored several pruning structures and DNN architectures which indicate different potential with respect to computation and memory efficiency.
This section evaluates how these metrics impact the inference speed in terms of throughput.
The same models are used as in Section~\ref{sec:pruning}, i.e., different WRN variants and DenseNet on the CIFAR-10 task.
Figure~\ref{fig:throughput_cifar} reports key inference metrics using the TensorRT framework targeting a Jetson Nano GPU.
Half-precision floating-point is used for weights as well as activations and other reduced precision formats are omitted because the accelerator and its software infrastructure do not support arbitrary precision types.

As can be seen, the various models in each regime (pruning structure or DNN architecture) show similar behavior for throughput.
The worst performing regimes are group and kernel pruning as well as the combination of fixed grouping and channel pruning. 
Especially interesting is group pruning: although it greatly reduces FLOPs and parameters, it fails at translating this reduction into faster computations. 
In contrast, employing pure channel pruning with residual and dense architectures yields the best performance. 
These results highlight the importance of reducing memory (or more specifically activations) rather than FLOPs in order to reduce latency or increase throughput.

\subsubsection{Overall Comparison}
\label{sec:overall}
\begin{figure}[t]
	\begin{center}
		\includegraphics[width=0.7\textwidth]{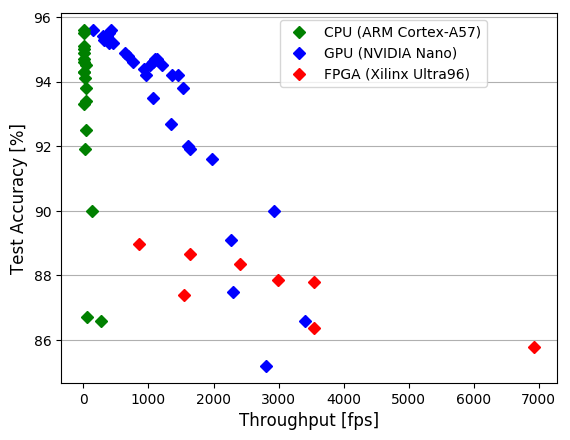}
		\caption{Throughput-accuracy trade-off of different compression methods for different processor architectures (CPU, FPGA, GPU) on the CIFAR-10 task.}
		\label{fig:throughput_final}
	\end{center}
\end{figure}

The previous sections studied several compression methods with considerations on the targeted software and hardware stack. 
Each section had a focus on leveraging compression to accelerate inference computations as much as possible while maintaining the prediction quality of the uncompressed model. 
In this section we compare these specialized forms of compression on their respective hardware in terms of absolute performance to identify the most promising compute concepts for DNNs.
Notably, whilst fundamentally different in architecture, from a system-level view these three processors, namely ARM Cortex-A57 CPU, NVIDIA Nano GPU, and XILINX Ultra96 FPGA, are comparable as they all exhibit a power consumption in the range of about 5 Watts.
Figure~\ref{fig:throughput_final} reports throughput and accuracy for these devices using different compression methods.

CPUs are well suited for mapping compressed DNNs.
However, the comparison to massively parallel processors shows that they lack the necessary amount of parallelism to achieve competitive throughput. 
The benefit of CPUs is that they feature a relatively large memory, allowing large and accurate models to be deployed.

FPGAs excel at extremely-high-throughput inference and high utilization of the available hardware resources. 
Such data-flow architectures, however, demand the entire model (including activations) to stay on chip, possibly preventing larger and more accurate models to be deployed.

GPUs are relatively constrained in terms of flexibility when deploying compressed models, due to the requirement of using optimized libraries and their respective software stack. 
However, they show a good compromise of programmable, general-purpose processing, enabling high throughput and accuracy. 
Additionally, they feature a large off-chip memory which in combination with latency hiding techniques enables high-throughput inference for large models.

\section{Conclusion} \label{sec:conclusion}
We presented an overview of the vast literature of the highly active research area concerned with resource efficiency of DNN inference.
We have identified three major directions of research, namely (i) network quantization, (ii) network pruning, and (iii) approaches that target efficiency at the structural level.
Many of the presented works are orthogonal and can in principle be used in combination to potentially further improve the results reported in the respective papers.

We have discovered several patterns in the individual strategies for enhancing resource efficiency.
For quantization approaches, a common pattern in the most successful approaches is to combine real-valued representations, that help in maintaining the expressiveness of DNNs, with quantization to enhance the computationally intensive operations.
More recently, mixed-precision quantization, where the bit widths are determined during training, is an upcoming topic.
For pruning methods, we observed that the trend is moving towards structured pruning approaches that obtain smaller models whose data structures are compatible with highly optimized dense tensor operations.
On the structural level of DNNs, a lot of progress has been made in the development of specific architectures that maintain a high expressiveness of the DNN while at the same time reducing the computational overhead substantially.
The newly emerging NAS approaches are promising candidates to automate the design of application-specific architectures with little user interaction.
However, it appears unlikely that current NAS approaches will discover new fundamental design principles as the resulting architectures highly depend on a-priori knowledge encoded in the architecture search space.

In experiments, we demonstrated on two benchmark data sets the difficulty of finding a good trade-off among prediction quality, representational efficiency and computational efficiency.
Considering three embedded hardware platforms, we showed that massive parallelism is required for inference efficiency and that quantization as well as structured pruning map well onto these accelerators.
We furthermore point out that hardware properties and the corresponding computational efficiency form a large fraction of resource efficiency.
This highlights the need to consider particular hardware targets when searching for resource-efficient machine learning models.

\section*{Acknowledgments}
This work was supported by the Austrian Science Fund (FWF) under the project number I2706-N31 and the German Research Foundation (DFG) under the project number FR3273/1-1. 
Furthermore, we acknowledge the LEAD Project Dependable Internet of Things funded by Graz University of Technology. 
This project has received funding from the European Union's Horizon 2020 research and innovation programme under the Marie Sk\l{}odowska-Curie Grant Agreement No.~797223 --- HYBSPN.
Also, this work is part of the COMET program within the K2 Center “Integrated Computational Material, Process and Product Engineering (IC-MPPE)” (Project No 886385), and supported by the Austrian Federal Ministries for Climate Action, Environment, Energy, Mobility, Innovation and Technology (BMK) and for Labour and Economy (BMAW), represented by the Austrian Research Promotion Agency (FFG), and the federal states of Styria, Upper Austria and Tyrol.
We acknowledge NVIDIA for providing GPU computing resources. 

\bibliography{jmlr2019_references}

\clearpage

\end{document}